\pdfoutput=1

\documentclass[11pt]{article}

\usepackage[table]{xcolor}

\usepackage[preprint]{acl}

\usepackage{times}
\usepackage{latexsym}

\usepackage[T1]{fontenc}

\usepackage[utf8]{inputenc}

\usepackage{microtype}

\usepackage{inconsolata}

\usepackage{graphicx}

\usepackage{booktabs}
\usepackage{array} 

\definecolor{lightgreen}{RGB}{200,255,200}
\definecolor{lightyellow}{RGB}{255,255,200}
\definecolor{lightred}{RGB}{255,200,200}
\definecolor{lightblue}{RGB}{200,230,255}
\definecolor{lightpurple}{RGB}{230,200,255}
\definecolor{lightorange}{RGB}{255,230,200}
\usepackage{amsmath}
\usepackage{multirow} 
\usepackage{subfigure}

\usepackage{fontawesome5}

\newcommand{\pipename}[0]{CoT2EL}

\title{Threading the Needle: Reweaving Chain-of-Thought Reasoning \\ to Explain Human Label Variation}

\author{
 \textbf{Beiduo Chen\textsuperscript{\faMountain\kern1pt\faRobot}} \quad
 \textbf{Yang Janet Liu\textsuperscript{\faBell}} \quad
 \textbf{Anna Korhonen\textsuperscript{\faSchool}} \quad
 \textbf{Barbara Plank\textsuperscript{\faMountain\kern1pt\faRobot}}
\\
\textsuperscript{\faMountain} MaiNLP, Center for Information and Language Processing, LMU Munich, Germany \\
\textsuperscript{\faRobot} Munich Center for Machine Learning (MCML), Munich, Germany \\
\textsuperscript{\faBell} Department of Linguistics, University of Pittsburgh, USA \\
\textsuperscript{\faSchool} Language Technology Lab, University of Cambridge, United Kingdom \\
\tt{ \href{mailto:beiduo.chen@lmu.de}{\textcolor{black}{beiduo.chen@lmu.de}},
\href{mailto:jal787}{\textcolor{black}{jal787@pitt.edu}}, 
\href{mailto:alk23@cam.ac.uk}{\textcolor{black}{alk23@cam.ac.uk}},
\href{mailto:b.plank@lmu.de}{\textcolor{black}{b.plank@lmu.de}}
}}

\begin{document}
\maketitle
\begin{abstract}

The recent rise of reasoning-tuned Large Language Models (LLMs)—which generate chains of thought (CoTs) before giving the final answer—has attracted significant attention and offers new opportunities for gaining insights into human label variation, which refers to plausible differences in how multiple annotators label the same data instance.
Prior work has shown that LLM-generated explanations can help align model predictions with human label distributions, but typically adopt a \textit{reverse} paradigm: producing explanations based on given answers. In contrast, CoTs provide a \textit{forward} reasoning path that may implicitly embed rationales for each answer option, before generating the answers. 
We thus propose a novel LLM-based pipeline enriched with linguistically-grounded discourse segmenters to extract supporting and opposing statements for each answer option from CoTs with improved accuracy. 
We also propose a rank-based HLV evaluation framework that prioritizes the ranking of answers over exact scores, which instead favor direct comparison of label distributions.
Our method outperforms a direct generation method as well as baselines on three datasets, and shows better alignment of ranking methods with humans, highlighting the effectiveness of our approach.

\end{abstract}

\newcolumntype{P}[1]{>{\raggedright\arraybackslash}p{#1}}

\section{Introduction}

\begin{figure}[t]

        \centering
        \includegraphics[width=\linewidth]{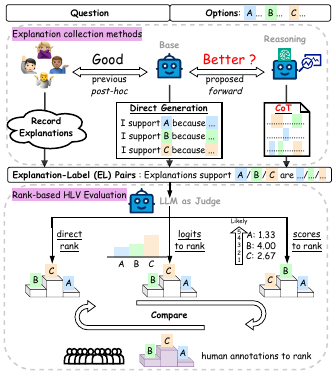}
        \caption{
        This motivational schematic illustrates three alternative paths—human annotation \cite{chen-etal-2024-seeing}, direct generation \cite{chen-etal-2025-rose}, and CoT2EL—that converge on the same EL-style output.
        We $\mathrm{i}$) repurpose the reasoning content in CoTs as forward and label-free method to extract  explanations for HLV, instead of direct generation (top); and $\mathrm{ii}$) propose a rank-based HLV evaluation framework (bottom).
       }
        \label{fig:figure1}
\end{figure}

Recent advances in large language models (LLMs, \citealt{DBLP:journals/corr/abs-2307-09288,DBLP:journals/corr/abs-2407-21783,DBLP:journals/corr/abs-2303-08774}) have shown the power of chain-of-thought (CoT, \citealt{DBLP:conf/nips/Wei0SBIXCLZ22,DBLP:conf/iclr/0002WSLCNCZ23}) reasoning in improving complex decision-making tasks \cite{wei2023chainofthoughtpromptingelicitsreasoning,DBLP:journals/corr/abs-2312-11562,DBLP:journals/csur/YuZTW24,DBLP:journals/corr/abs-2501-12599,DBLP:journals/corr/abs-2503-09567}. One prominent direction involves reasoning-tuned LLMs, which generate CoT reasoning steps explicitly before producing a final answer, often guided by reinforcement learning to promote interpretable and structured thinking processes~\cite{DBLP:journals/corr/abs-2501-12948,qwq32b,DBLP:journals/corr/abs-2410-21276}. While prior work has primarily focused on analyzing the content and structure of CoTs to improve accuracy or interpretability~\cite{DBLP:journals/corr/abs-2410-18982,DBLP:journals/corr/abs-2412-09413,ameisen2025circuit}, little attention has been given to \emph{the potential of CoTs in capturing more nuanced aspects of human annotation behavior}. In particular, human label variation (HLV, \citealt{plank-2022-problem}) arises when different annotators provide divergent yet valid labels for the same input, a phenomenon especially common in inference and multiple-choice tasks involving ambiguous, subjective, or commonsense-rich questions \cite{pavlick-kwiatkowski-2019-inherent,DBLP:journals/aim/AroyoW15}. Modeling HLV is thus crucial for creating robust NLP systems that reflect the diversity of human perspectives~\cite{DBLP:journals/jair/UmaFHPPP21,plank-2022-problem}.

Prior research has shown that explanation-label pairs—either produced by humans or models—can help LLMs better capture the distribution of human labels~\cite{weber-genzel-etal-2024-varierr,chen-etal-2025-rose,chen-etal-2024-seeing}. However, existing approaches treat model explanation generation as a \textit{post-hoc} task, generating explanations after a label is chosen~\cite{chen-etal-2025-rose}. In contrast, reasoning-tuned LLMs offer a \textbf{\textit{forward}} reasoning paradigm: CoTs precede answer selection and may already contain latent rationales for why certain labels are chosen—rationales that, if properly extracted, could serve as label-specific explanations.

In this work, we investigate \textbf{whether CoTs can be repurposed as a source to extract explanation-label pairs to derive insights on HLV}, as visualized in Figure \ref{fig:figure1}. 
Specifically, we propose a novel pipeline, {\pipename}, that includes discourse segmenters to extract such pairs from CoTs. 
Such an approach allows us to view CoTs not merely as reasoning artifacts, but as explanation-rich representations that reflect a broader label space.

\begin{figure}[t]
    \centering
    \subfigure[rank versus value]{
        \includegraphics[width=0.23\textwidth]{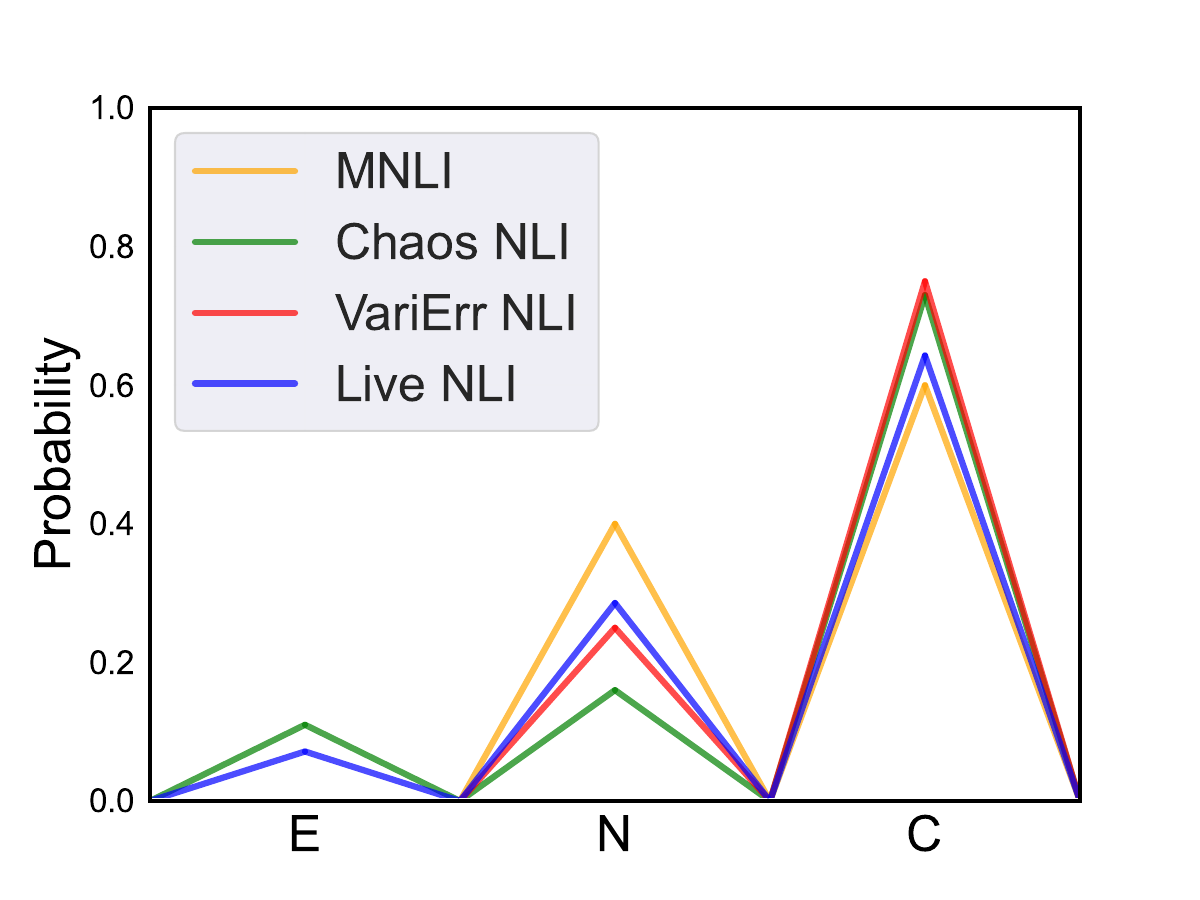}\label{fig:ex_comparison}
    }
    \subfigure[closed-world assumption]{
        \includegraphics[width=0.22\textwidth]{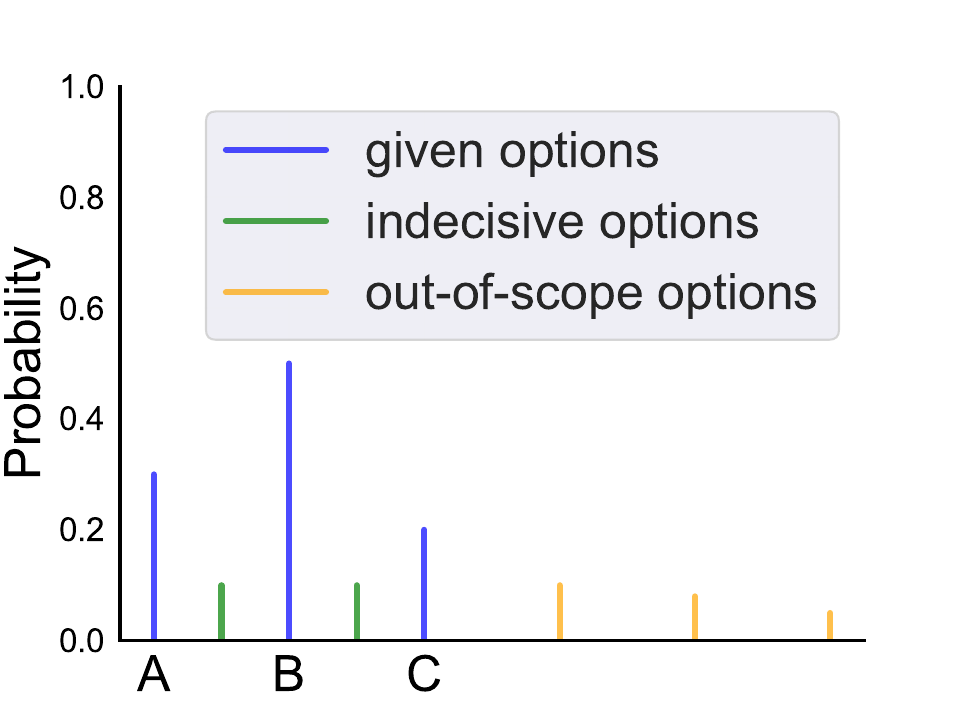}\label{fig:ex_nonexclusive}
    }
    \caption{(a) Same instance, different NLI datasets: probability  values differ, rank the same. The y-axis is human label selection probability. (b) is a conceptual illustration to introduce human answer behavior across MCQA datasets compared to the enforced closed-world assumption in normalized label probabilities.}
    \label{fig:evaluation_motivation}
\end{figure}

We further propose a new HLV evaluation framework centered around \textbf{\textit{ranking}} rather than label distributions. Current HLV evaluations assume closed-label sets and primarily focus on approximating exact probability distributions.
However, exact value differences may only matter if they yield differences in label preferences (rankings), cf.\ Figure~\ref{fig:ex_comparison}. They can also be highly sensitive to annotator variability and availability, and a closed-set,  i.e.\ Figure~\ref{fig:ex_nonexclusive}, limits their ability to capture broader possibilities.
Therefore, we evaluate how well model-predicted rankings over options align with human rankings, providing a more nuanced and robust view of model performance in settings where annotation disagreements exist.

We conduct extensive experiments on three benchmarks exhibiting label variation: VariErr NLI \cite{weber-genzel-etal-2024-varierr}, CommonsenseQA \cite{talmor-etal-2019-commonsenseqa}, and Social IQa \cite{sap-etal-2019-social}. 
Our results across multiple LLM judges demonstrate that explanation-label pairs extracted from CoTs using our {\pipename} pipeline consistently outperform both the direct explanation generation method and explanation-free baselines in capturing annotation disagreements. 
Our analysis and ablation studies further reveal several key insights: $\mathrm{i}$) \textit{supporting} statements within CoTs are more predictive of human choices than the opposing ones; $\mathrm{ii}$) ranking strategies aligned with human annotation practices lead to better approximation of human label variation; and $\mathrm{iii}$) well-structured input prompts significantly enhance the informativeness of CoTs.

\newcommand{\hlc}[2]{\setlength{\fboxsep}{1pt}\colorbox{#1}{#2}}

\begin{table*}[t]
\centering
\resizebox{\textwidth}{!}{\scriptsize
\begin{tabular}{P{\textwidth}}
\toprule 
\textbf{Input:} Please read the following social scenario and the accompanying question, choose the most appropriate answer from the options provided and start your answer with a single letter. \, Scenario: Ash redeemed themselves after retaking the test they failed. \, Question: How will Ash feel as a result? \newline A. relieved \, B. accomplished \, C. proud \, Answer: \\
\midrule
\textbf{CoTs:} Okay, let's see. The scenario is that Ash redeemed themselves by retaking a test they failed. The question is asking how Ash will feel as a result. The options are \hlc{lightblue}{A. relieved}, \hlc{lightgreen}{B. accomplished}, and \hlc{lightorange}{C. proud}.
\newline First, I need to understand what each of these emotions means in this context. Redeeming oneself after failing a test suggests that Ash probably worked hard to retake it and succeeded this time. So, the feelings here would be related to overcoming a challenge.
\newline "Relieved"(A) would mean that Ash is feeling a release from the stress or worry about having failed before. \hlc{lightblue}{Passing the test might take away that anxiety.}
\newline "Accomplished"(B) implies a sense of achievement, having completed something successfully, especially after effort. \hlc{lightgreen}{Since Ash had to retake the test, putting in the} \newline \hlc{lightgreen}{work to pass it now would make them feel like they've accomplished something.}
\newline "Proud"(C) is more about self-esteem or satisfaction in one's own achievements. \hlc{lightorange}{It's a bit similar to accomplished but more focused on the personal pride aspect.}
\newline Now, considering the scenario: Ash didn't just pass a test; they redeemed themselves. \hlc{lightblue}{That suggests that failing initially might have caused some shame or} \newline \hlc{lightblue}{disappointment, so passing now would alleviate that.}\hlc{lightgreen}{But redemption often involves not just relief but also a sense of achievement.}
So both A and B are possible. 
\newline But "proud" might be more about the internal sense of pride, whereas "accomplished" is about the feeling of having achieved something through effort. Which one fits better here? \hlc{lightgreen}{If Ash worked hard to retake and pass, then feeling accomplished makes sense.} Relieved is possible, but maybe the primary feeling is accomplishment. \hlc{lightorange}{Proud could also be there,} but maybe \hlc{lightgreen}{B is more directly related to the effort and success.}
\newline I think the best answer is B. Accomplished.
\\ 
\bottomrule
\end{tabular}}
\caption{
\textbf{CoT Example}. The examples, drawn from Social IQa, are processed using DeepSeek R1~\cite{DBLP:journals/corr/abs-2501-12948}, with CoT segments related to options A, B, and C highlighted in blue, green, and orange, respectively. For questions where all options appear plausible, the CoT reflects a step-by-step analysis involving comparison and evidence evaluation. These segments offer rich material for deriving explanations relevant to our study of the HLV phenomenon, enabling a deeper understanding of the task and the roots of HLV beyond just the final answers.
}
\label{tab:CoT_case}
\end{table*}

\section{Background and Motivation}

\subsection{Modeling Human Label Variation}
\label{subsec:motivation1}

Most current approaches to investigating HLV view the label distribution from annotators as a probability distribution (e.g.,~\citealp[]{DBLP:journals/corr/abs-2502-01891,DBLP:journals/jair/UmaFHPPP21,pavlick-kwiatkowski-2019-inherent,nie-etal-2020-learn,lee-etal-2023-large,leonardelli-etal-2023-semeval,rizzi-etal-2024-soft,pavlovic-poesio-2024-effectiveness}) or assign scores (e.g., \citealp[]{palta-etal-2024-plausibly,DBLP:journals/corr/abs-2305-14770}) to capture annotation disagreement. Evaluation typically focuses on how well models approximate these distributions or scores. However, we argue that such paradigms have key limitations:

\paragraph{Rank or Value: Two Complementary Views}
Human annotations inherently reflect subjective judgments, and when different groups of annotators are involved, the resulting label distributions can vary significantly. To illustrate this variation, we examine four NLI datasets annotated by distinct human populations: MNLI~\cite{williams-etal-2018-broad}, Chaos NLI~\cite{nie-etal-2020-learn}, VariErr NLI~\cite{weber-genzel-etal-2024-varierr}, and Live NLI~\cite{jiang-etal-2023-ecologically}. Among $15$ shared instances across these datasets, only $2.2$\% of pairwise comparisons yield identical probability values, underscoring the sensitivity of distributional scores to annotator composition. Despite these numeric differences, $43.3$\% of the comparisons retain consistent label rankings (Figure~\ref{fig:ex_comparison}). This suggests that while probability-based evaluations are highly unstable under annotator shifts, rank-based assessments exhibit greater consistency. Given that many real-world applications rely more on the correct ordering of label plausibility than exact probability estimates, we argue for the inclusion of \textbf{\textit{ranking}} as a complementary and more robust HLV evaluation metric.

\paragraph{Does the Closed-World Assumption Reflect Human Judgments?}
Human annotators do not always treat label options as strictly exhaustive or mutually exclusive. In some contexts, they express uncertainty or endorse multiple plausible answers, often via indecisive choices 
as observed in datasets like Live NLI and VariErr NLI (green lines in Figure~\ref{fig:ex_nonexclusive}). Converting such annotations into normalized probability distributions imposes a closed-world assumption—requiring mutually exclusive, collectively exhaustive labels summing to one (blue lines)—which limits the label space and overlooks ambiguous or open-ended responses common in tasks like CommonsenseQA \cite{talmor-etal-2019-commonsenseqa} or Social IQa~\cite{sap-etal-2019-social} (orange lines). 
This constraint can distort model evaluation by masking ambiguity.
We therefore propose a rank-based evaluation framework (\S\ref{sec:rank-based-eval}), which better accommodates indecisive and out-of-scope options.

\subsection{Modeling HLV with Explanations}
\label{subsec:motivation2}

Recent studies have shown that explanations can effectively support the interpretation and analysis of HLV~\cite{DBLP:journals/corr/abs-2304-12443,chen-etal-2024-seeing,weber-genzel-etal-2024-varierr,jiang-etal-2023-ecologically}. 
However, collecting human explanations is significantly more resource-intensive than traditional label-only annotation.
To reduce annotation costs, recent studies have leveraged LLMs to generate explanations for each label. Evidence shows that with a few human labels, LLM-generated explanations can rival human-written ones in forming valid explanation-label pairs and supporting HLV modeling~\cite{chen-etal-2025-rose}. However, this approach has three key limitations: $\mathrm{i}$) it relies on a few human labels to select final explanations, with performance degrading when such supervision is absent; $\mathrm{ii}$) it reverses the annotation process by conditioning explanation generation on labels, risking hallucinated reasoning for implausible options; and $\mathrm{iii}$) it treats labels independently, lacking comparative reasoning and thus reducing explanation depth and completeness.
To address these limitations, we study how the potential of CoTs (Table~\ref{tab:CoT_case}) from LLMs can be leveraged to explain HLV, given their rich argumentations and consideration of multiple alternative options.

\section{Datasets}
\label{sec:datasets}

To study HLV via explanation-based methods, we select datasets with multiple annotation choices. 
An overview of the selected datasets are shown in Table~\ref{tab:datasets}. 
Specifically, \textbf{VariErr NLI}~\cite{weber-genzel-etal-2024-varierr} is a Natural Language Inference (NLI) dataset which includes annotations and human-provided explanations from four annotators. Notably, there are $500$ NLI instances that also overlap with the Chaos NLI and MNLI datasets, providing label distributions from $100$ and five annotators for each instance, respectively. This makes VariErr NLI an especially valuable dataset for conducting rational, explanation-based analysis of annotation disagreement in inference tasks. 

In addition, we include two multiple-choice question answering (MCQA) datasets: Social IQa (\textbf{SIQA},~\citealt{sap-etal-2019-social}) and CommonsenseQA (\textbf{CQA},~\citealt{talmor-etal-2019-commonsenseqa}). Both require general world knowledge and reasoning to answer correctly. Importantly, \citet{palta-etal-2024-plausibly} re-annotated these datasets, collecting Likert-scale ratings (from $1$ to $5$) from five annotators for each answer option~\cite{zhang-etal-2017-ordinal}, as well as human feedback for hard-to-judge items. The mean rating is then used as the option's plausibility score. This approach offers a new angle for studying HLV.

\begin{figure*}[t!]

        \centering
        \includegraphics[width=\linewidth]{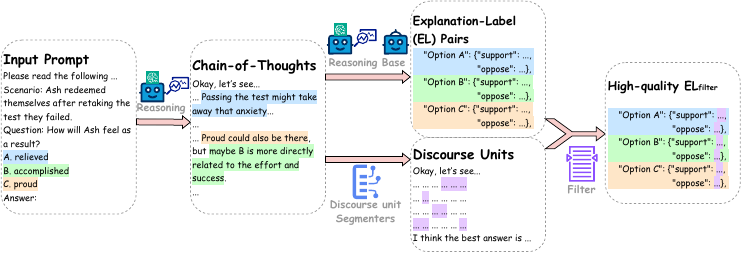}
        \caption{Overall structure of the proposed explanation-label (EL) pair extraction pipeline. Details in Appendix~\ref{app:pipeline-prompt}.}
        \label{fig:pipeline}
\end{figure*}

\begin{figure*}[t!]

        \centering
        \includegraphics[width=\linewidth]{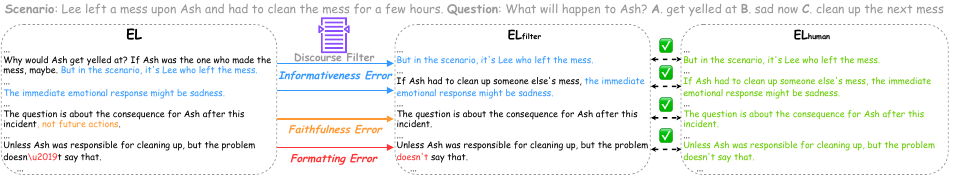}
        \caption{Three error types in ELs during LLM parsing (left) and the human validation procedure (right).}
        \label{fig:human_validation}
\end{figure*}

\section{Extracting Explanation-Label Pairs from Chain-of-Thought Reasoning}
\label{sec:extraction_pipeline}

CoT reasoning provides rich rationales (Table~\ref{tab:CoT_case}) to support decision-making in tasks like MCQA. 
However, extracting fine-grained, option-specific explanations from CoTs is non-trivial due to the lack of explicit alignment between reasoning fragments and individual answer options. Below we describe our proposed method for extracting and refining structured explanation-label ({EL}) pairs from CoTs using LLMs as parsers and two linguistic-motivated discourse unit segmenters.

\begin{table}[t!]
\centering
\resizebox{\linewidth}{!}{
\begin{tabular}{P{0.20\textwidth} |P{0.20\textwidth}|P{0.20\textwidth}}
\toprule 
\multicolumn{1}{l|}{\textbf{HLV Datasets (num.)}}  & \multicolumn{1}{l|}{\textbf{Instance Content}} & \multicolumn{1}{l}{\textbf{Annotations}} \\
\midrule
\textbf{VariErr NLI} \newline ($500$) \newline \textit{from MNLI dev set}  & hypothesis \newline premise \newline $3$ NLI labels (ENC) & $4$ from VariErr NLI \newline $100$ from Chaos NLI \newline $5$ from MNLI  \\ \midrule
\textbf{Social IQa} \newline ($125$) \newline \textit{from SIQA dev set}
& social scenario \newline question \newline $3$ options (ABC) & $5$ annotators score one question-option pair individually.  \\ \midrule
\textbf{CommensenseQA} \newline ($125$) \newline \textit{from CQA dev set} & question \newline $5$ options (ABCDE) & $5$ annotators score one question-option pair individually.  \\ 
\bottomrule
\end{tabular}}
\caption{An Overview of the Datasets.}
\label{tab:datasets}
\end{table}

\subsection{{\pipename} Pipeline}

Our method is designed to produce a set of EL pairs that represent supporting or opposing arguments for each answer option in a given MCQA task. The full pipeline is depicted in Figure~\ref{fig:pipeline}.\!\footnote{Code is available at \href{https://github.com/mainlp/CoT2EL}{https://github.com/mainlp/CoT2EL}.}

\paragraph{CoT Generation and Initial Extraction.}
Given a question \( Q \) and a set of candidate labels \( \mathbf{L} = [l_1, l_2, ..., l_n] \), we first prompt a reasoning-tuned LLM to generate a CoT reasoning:

{\small
\begin{equation}
\text{CoT} = \text{ReasoningModel}(Q, \mathbf{L}),
\label{eq:cot_generation}
\end{equation}}

\noindent We then apply both a reasoning-tuned model and its corresponding base model in sequence, which is used as a structured output parser.
Specifically, the CoT content is converted into a structured \texttt{JSON} list of {EL} pairs in the following format:

{\small
\begin{equation}
\left\{(e_i, l_x, s_i) \;\middle|\; e_i \in \text{CoT},\ l_x \in \mathbf{L},\ s_i \in \{\texttt{support}, \texttt{oppose}\} \right\}
\label{eq:structured_el}
\end{equation}
}

\noindent where $e$ is the full generated explanation extracted from CoT, and $s_i$ indicates whether the explanation $e_i$ supports or opposes the label $l_i$.
While LLM-based parsers are able to parse and decompose the CoT content, the directly extracted {EL} pairs often exhibit issues regarding---which we categorize into the following aspects (exemplified in Figure~\ref{fig:human_validation}):
$\mathrm{i}$) \textbf{Informativeness}: explanations either lack key content or contain unnecessary information.
$\mathrm{ii}$) \textbf{Faithfulness}: statements may paraphrase or hallucinate beyond the original CoT content.
$\mathrm{iii}$) \textbf{Formatting}: structural inconsistencies or unexpected formatting issues arise in the generated \texttt{JSON} outputs.
These issues complicate the direct use of such pairs for downstream reasoning evaluation and necessitate further refinement steps.

\paragraph{Discourse-guided Refinement.}
To mitigate the aforementioned issues, we apply two discourse segmenters ($\text{DSeg}_{{i}}$) 
that offer complementary views of text structure to segment the CoT content into a set of coherent discourse units:
a discourse unit segmenter following the Rhetorical Structure Theory (RST, \citealt{mann1988rhetorical}), which segment sentences into clause-based units;
a discourse connective detector following the Penn Discourse Treebank (PDTB, \citealt{webber2019penn}), which identifies clauses initiated with connectives (e.g.~\textit{however}, \textit{because}) that signal relationships between ideas. See Figure \ref{fig:human_validation} for filtered instances (in the $\text{EL}_{\text{filter}}$ box). 
By combining both, we obtain a richer set of candidate segments. This enhances the precision of filtering noisy or mismatched EL pairs generated by the reasoning LLMs, making the extracted explanation structure more accurate and interpretable.

The integration of discourse segmentation into our pipeline is driven by the necessity to extract logically coherent and interpretable reasoning units from CoT content. In human annotation practices, such units often form the basis for identifying justifications that support or oppose specific answer choices. By emulating this annotation logic through automated discourse models, we impose structural and semantic regularity on the extracted explanations. This approach facilitates reliable interpretation and alignment of explanation-label pairs, thereby enhancing the transparency and evaluability of CoT reasoning.
Both discourse segmenters are trained using the DISRPT Shared Task winning system DisCoDisCo \cite{gessler-etal-2021-discodisco} with the DISRPT 2023 Shared Task data \cite{braud-etal-2023-disrpt} (see Appendix~\ref{app:discodisco-training-results} for training and performance details). The outputs are then processed and merged into a unified set of valid semantic discourse units:

{\small
\begin{equation}
U = \text{DSeg}_1(\text{CoT}) \cup \text{DSeg}_2(\text{CoT}),
\label{eq:semantic_units}
\end{equation}

}

This normalized set \( U \) forms a constrained, high-quality space of candidate explanation units, grounded directly in the original CoT.
We align each extracted explanation \( e_i \) from Eq~\ref{eq:structured_el} with its closest discourse unit in \( U \) by maximal similarity:\footnote {Implemented by Python \texttt{difflib.SequenceMatcher}.}

{\small
\begin{equation}
{
\text{EL}_{\text{filter}} = \left\{(e^*_i, l_x, s_i) \;\middle|\; e^*_i = \arg\max_{u \in U} \text{Sim}(u, e_i) \right\}.
}
\label{eq:final_el}
\end{equation}
}

The final result \( \text{EL}_{\text{filter}} \) is a set of {EL} pairs in which each explanation is both semantically faithful and textually aligned with a coherent discourse unit from the original CoT content. This structured output enhances both interpretability and utility for evaluating reasoning processes in MCQA settings.

\subsection{Validation through Human Annotation}
\label{sec:human_eval}

To assess the reliability and effectiveness of our pipeline, {\pipename}, we conducted a human annotation study across the three datasets. 
We randomly sampled $10$ CoT instances from each dataset. For each instance, a trained annotator\footnote{The annotator is paid according to national standards.} manually identified and labeled all explanation spans within the CoT content that either supports or opposes a given answer label, using the target format illustrated in Figure~\ref{fig:pipeline}. This produced a human-curated gold standard of EL pairs for comparison.

DeepSeek R1 660B~\cite{DBLP:journals/corr/abs-2501-12948} was used to generate CoT responses. The corresponding base model, DeepSeek V3~\cite{DBLP:journals/corr/abs-2412-19437}, was then incorporated to standardize the CoT into structured {EL} via \texttt{JSON} parsing.
Following the pipeline in Figure~\ref{fig:pipeline}, we applied the two discourse segmenters to produce the final \(\text{EL}_{\text{filter}}\). As shown in Figure~\ref{fig:human_validation}, these auto-generated pairs were then quantitatively compared to the human-annotated counterparts across four evaluation dimensions:
lexical, syntactic, semantic similarities, and {Levenshtein ratio} \cite{giulianelli-etal-2023-comes}.
Besides evaluating the full {EL} sets, we also considered the supporting-only settings (EL-sup and \(\text{EL}_{\text{filter}}\text{-sup}\)), which aligns with the direct LLM generation method that favors positive justifications.

\begin{table}[th]
\centering
\resizebox{0.8\linewidth}{!}{
\begin{tabular}{lccc}
\toprule
\textbf{Datasets} & \textbf{VariErr NLI} & \textbf{SIQA} & \textbf{CQA} \\
\midrule
\, \(\text{EL}\) & 0,6820 & 0,7897 & 0,8200 \\
\, \(\text{EL}_{\text{filter}}\) & 0,8106 &0,8761& 0,8684 \\
\, \(\text{EL-sup}\) &  0,6992 & 0,8167 & 0,8431 \\
\, \(\text{EL}_{\text{filter}}\text{-sup}\)  &{0,8296} & {0,8825}&0,8749 \\
\bottomrule
\end{tabular}}
\caption{Averaged scores among 4 metrics (Lexical, Syntactic, Semantic Similarities and {Levenshtein Ratio}) for human validation. Higher score, more similar.}
\label{tab:validation-avg}
\end{table}

The comparison in Table~\ref{tab:validation-avg} shows that our final set \(\text{EL}_{\text{filter}}\) more closely aligns with human annotations than unfiltered EL.\!\footnote{Detailed metrics and scores are in Appendix~\ref{app:human-anno-eval-dimensions}.}
This suggests that our discourse-guided extraction pipeline achieves a high degree of faithfulness and interpretability, approximating human performance in identifying rationale-label mappings from CoT content.

\section{Rank-based HLV Evaluation}
\label{sec:rank-based-eval}

Recent studies employ the \textit{LLM-as-judge} paradigm~\cite{DBLP:conf/nips/ZhengC00WZL0LXZ23}, wherein explanations accompany questions and candidate labels as inputs to an LLM\footnote{We elaborate the prompt for injection of EL pairs into LLM judges in Table~\ref{tab:app4-EL-injection_prompt} in Appendix~\ref{app:ranking-method}.}~\cite{chen-etal-2025-rose,chen-etal-2024-seeing}. The resulting output distribution is evaluated against the empirical human label distribution, using alignment as a proxy for explanation quality.

We propose a rank-based evaluation framework as a more robust complement to raw probability comparisons. Building on the \textit{LLM-as-judge} paradigm, our approach shifts the evaluation focus to label ranking. Human annotations from HLV datasets are transformed into rankings, and the LLM is prompted to generate corresponding rankings based on the input of questions and options. Model-generated rankings are then compared to human-derived rankings as explanation-free baselines. To assess the impact of explanations, we additionally provide EL pairs and evaluate whether they enhance alignment with human rankings.

\subsection{Ranking Generation Methods}
\label{subsec:ranking-gen-methods}

We experiment three distinct approaches to eliciting label rankings from LLMs:\footnote{Details in Appendix~\ref{app:ranking-method}.}

\noindent $\mathrm{i}$) \textbf{Direct Ranking} (\texttt{Rank-rank}): an LLM is explicitly instructed to rank the candidate labels based on the provided question, yielding a direct ranking. 

\noindent $\mathrm{ii}$) \textbf{First-Token-Logits Ranking} (\texttt{Rank-logits}): following prior work~\cite{DBLP:conf/icml/SanturkarDLLLH23,DBLP:journals/corr/abs-2306-16388, DBLP:journals/tmlr/LiangBLTSYZNWKN23}, the model is given a set of label options (A, B, C…) and asked to choose one. We then take the logits of the first output token for each label and use them to rank the labels from most to least likely.
This method produces a probability-like distribution by normalizing the logits over labels and is particularly designed to align with the distribution-based VariErr NLI. 

\noindent $\mathrm{iii}$) \textbf{Scoring-Based Ranking} (\texttt{Rank-score}): inspired by~\citet{palta-etal-2024-plausibly}, we prompt an LLM to assign each label a score from $1$ to $5$ based on its plausibility. The final ranking is derived from their scores.
This method is especially motivated by score-based SIQA and CQA.

\subsection{Evaluation Metrics}
\label{subsec:all_metrics_for_hlv}

To compare LLM rankings with humans, we compute two standard rank correlation metrics:
Kendall’s $\tau$ \cite{kendall1938new} and Spearman’s $\rho$ rank correlation coefficient~\cite{spearman1961proof}. 
Specifically, we assess three ranking generation methods proposed in \S\ref{subsec:ranking-gen-methods}: \texttt{Rank-rank}, \texttt{Rank-logits}, and \texttt{Rank-score}. 
We further compute appropriate similarity metrics to compare distributions from \texttt{Rank-logits} and scalar scores from \texttt{Rank-score} with human annotations. 
For probability distributions (from VariErr NLI), we use Kullback-Leibler (KL) Divergence \cite{kullback1951information}, Jensen-Shannon Distance (JSD, \citealt{DBLP:journals/tit/EndresS03}), and Total Variation Distance (TVD, \citealt{DBLP:books/daglib/0018090}). For scalar scores (from SIQA and CQA), we employ Root Mean Squared Error (RMSE, \citealt{hyndman2006another}), Mean Absolute Error (MAE, \citealt{willmott2005advantages}), and Coefficient of Determination ($R^2$, \citealt{steel1960principles}).
See details in Appendix~\ref{app:HLV-metric}.

\subsection{LLMs}
To generate CoTs, we used two reasoning-tuned LLMs: DeepSeek R1 660B (R1, ~\citealt{DBLP:journals/corr/abs-2501-12948}) and QwQ 32B (QwQ, ~\citealt{qwq32b}). For comparison with the direct explanation generation method, we additionally included their corresponding base LLMs: DeepSeek V3 (V3, ~\citealt{DBLP:journals/corr/abs-2412-19437}) and Qwen 2.5 Max (Qwen Max, ~\citealt{qwen2.5}).
For \textit{LLM-as-judge}, we adopt Qwen2.5-7B-Instruct (qwen,~\citealt{qwen2.5_normal,qwen2}), Llama-3.1-8B-Instruct (llama,~\citealt{DBLP:journals/corr/abs-2407-21783}), and Mistral-7B-Instruct-v0.2 (mistral,~\citealt{DBLP:journals/corr/abs-2310-06825}).

\begin{figure*}[t]

        \centering
        \includegraphics[width=\linewidth]{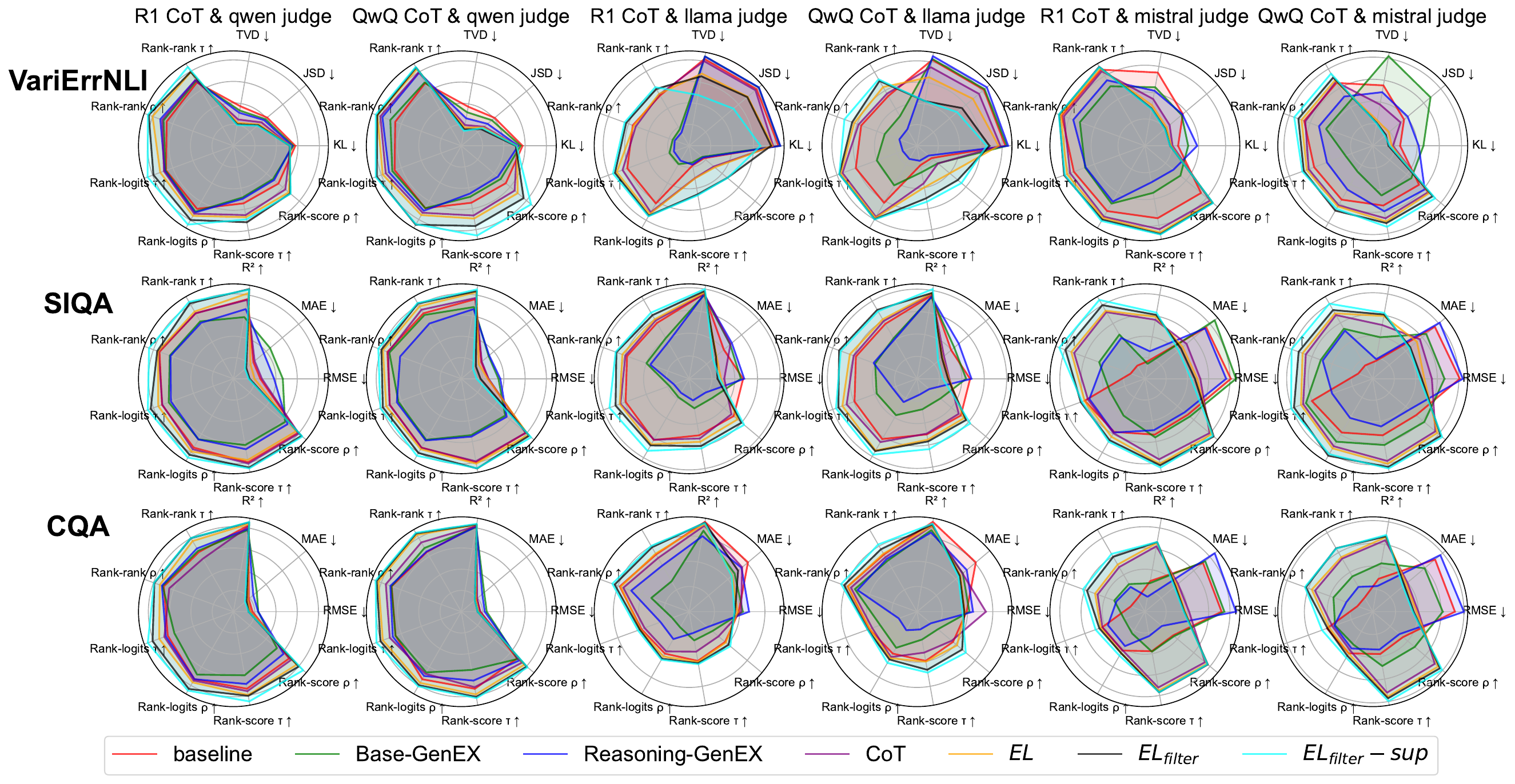}
        \caption{
\textbf{Radar charts present main results across datasets and settings.} Each chart spans nine axes, each representing a distinct evaluation metric, with arrows denoting the preferred performance direction. Columns correspond to evaluation settings wherein CoTs are generated by either R1 or QwQ and assessed by LLM judges including qwen, llama, and mistral. The red contour indicates the explanation-free baseline.
We evaluate various EL construction methods, including direct generation from reasoning-tuned or base LLMs (Reasoning/Base-GenEX), unprocessed CoT outputs, structured EL (Eq.\ref{eq:structured_el}), filtered outputs  \(\text{EL}_{\text{filter}}\) (Eq.\ref{eq:final_el}), and support-only content \(\text{EL}_{\text{filter}}\text{-sup}\). 
}
        \label{fig:main_results}
\end{figure*}

\section{Results and Analyses}
\label{sec:results}

Figure~\ref{fig:main_results} presents the main HLV evaluation results. 
Across nearly all metrics and settings, \(\text{EL}_{\text{filter}}\) and \(\text{EL}_{\text{filter}}\text{-sup}\) consistently achieve superior performance, outperforming both the explanation-free baseline and the direct generation method\footnote{Following~\citet{chen-etal-2025-rose}, the direct generation method includes Base-GenEX and Reasoning-GenEX using a base LLM and a reasoning-tuned LLM, respectively. Detailed implementation is in Appendix~\ref{app:pipeline-prompt}.} (GenEX), underscoring the effectiveness of the proposed {\pipename} pipeline in facilitating deeper HLV understanding and explaining.
Notably, although both only contain supporting rationals, \(\text{EL}_{\text{filter}}\text{-sup}\) yields a marked advantage over GenEX,
indicating that \textbf{the \textit{forward} paradigm and attention to inter-label dynamics enable reasoning-tuned models to generate CoTs with richer and more HLV-relevant content}, as motivated in \S\ref{subsec:motivation2}. 
Lastly, the consistent performance of our rank-based evaluation across both distributional and score-based settings affirms the robustness and generalizability of the proposed evaluation framework, as postulated in \S\ref{subsec:motivation1}.
The full results are in Appendix~\ref{app:all_results}.

\begin{figure*}[t] 
    \centering
    \includegraphics[width=\linewidth]{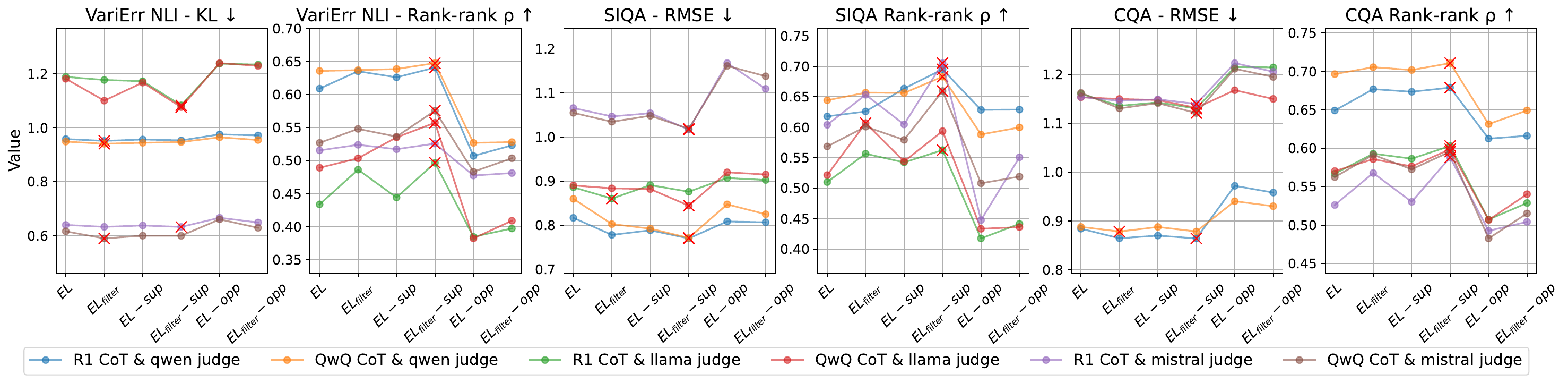}
    \caption{\textbf{Results of the ablation study.} The plots show the comparison of the effectiveness of \textit{supporting} versus \textit{opposing} explanations for HLV evaluation. Red crosses mark the best-performing data point for each setting. 
    }
    \label{fig:os_results}
\end{figure*}

\begin{figure}[t]
        \centering
        \includegraphics[width=\linewidth]{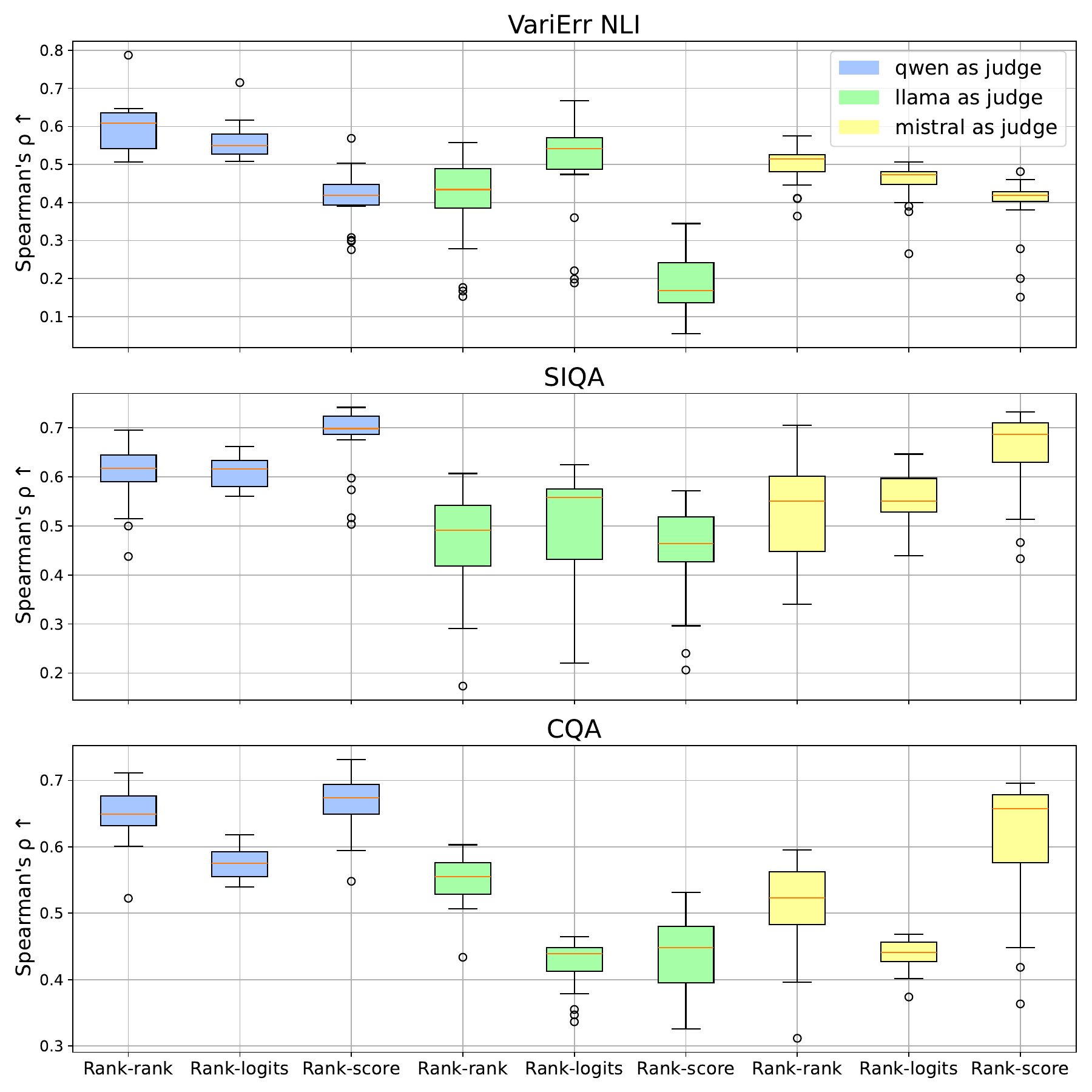}
        \caption{Comparison of three ranking generation methods across datasets and LLM judges. Each box represents the aggregated statistics of a given ranking method. 
        }
        \label{fig:rank_results}
\end{figure}

\paragraph{Support or Oppose?}
It is also worth noting that \(\text{EL}_{\text{filter}}\text{-sup}\) consistently outperforms \(\text{EL}_{\text{filter}}\), prompting further investigation into the effectiveness of supporting versus opposing explanations. We conducted an ablation study by isolating only the supporting and opposing components from both {EL} and \(\text{EL}_{\text{filter}}\), as shown in Figure~\ref{fig:os_results}.

It is clear that \(\text{EL}_{\text{filter}}\text{-sup}\) achieves the best results in most settings across all datasets,
while opposing-only explanations lead to performance degradation. A closer examination of individual {EL} pairs reveals two likely reasons for this outcome: $\mathrm{i}$) when rejecting a label, many CoTs tend to provide vague or ambiguous statements, whereas supporting statements for a label are often more affirmative and explicit; $\mathrm{ii}$) LLMs used as judges may be more influenced by the clearly articulated supporting reasoning. This ablation study not only reveals that support-oriented explanations are more effective for HLV modeling than oppose-oriented ones but also underscores the importance of training future LLMs to articulate rejections with greater clarity and confidence, rather than ambiguity.

\paragraph{How to rank?}
Our rank-based HLV evaluation framework applies three methods to obtain rankings from LLM judges, as detailed in \S\ref{subsec:ranking-gen-methods}.
We therefore investigate which ranking method yields the best performance, and present results in Figure~\ref{fig:rank_results}.
The comparison reveals several key patterns. For the distribution-based VariErr NLI, \texttt{Rank-logits} and \texttt{Rank-rank} achieve comparable average performance, whereas \texttt{Rank-score} performs consistently worse across all LLM judges. Conversely, in the score-based SIQA and CQA, \texttt{Rank-score} tends to outperform \texttt{Rank-logits}, aligning better with the annotation procedure. \texttt{Rank-rank}, the method in which the LLM judges directly rank the options, exhibits stable and competitive performance across all datasets and judges.

These findings confirm our motivations in ~\S\ref{subsec:ranking-gen-methods}, suggesting that the choice of the ranking method should ideally align with the annotation format used to construct the target HLV dataset—i.e., distributional versus score-based. Moreover, the robustness of \texttt{Rank-rank} highlights its general applicability across different HLV evaluations.

\paragraph{Structure Matters?}
We conducted an additional evaluation for the intermediate outputs in the {\pipename} pipeline, as shown in Table~\ref{tab:structure_results}. 
\(\text{CoT}_{\text{parser}}\) refers to the raw, unstructured explanations extracted from CoT by reasoning LLMs, before they are 
decomposed
into a strict \texttt{JSON} format to EL by base LLMs. 
We found that when the explanations are strictly structured, the LLM judge performs better than when using the original, unstructured ones. This shows that LLM judges utilize explanations more effectively when they are well-organized and explicitly indicate which parts support or oppose each answer choice.

We further analyze human explanations (HumanEX) across datasets as described in \S\ref{sec:datasets}, with a focus on how structural properties influence performance. In VariErr NLI, most instances provide $3$ to $6$ clear and high-quality human explanations that directly support specific answers, allowing for precise explanation-label pairs construction. 
In contrast, datasets like SIQA and CQA contain fewer and more vague human feedback, which are often only loosely marked as relevant.
Table~\ref{tab:structure_results} shows HumanEX performs significantly better on VariErr NLI than on SIQA or CQA—while explanation quality plays a key role, this also indirectly highlights the effectiveness of structured explanations.

\begin{table*}[t]
\centering
\resizebox{\linewidth}{!}{
\begin{tabular}{lccccccccccccccc}
\toprule
Datasets                          & \multicolumn{5}{c}{VariErr NLI}                                                                                                                             & \multicolumn{5}{c}{SIQA}                                                                                                                                     & \multicolumn{5}{c}{CQA}                                                                                                                                      \\ 
\cmidrule(lr){1-1} \cmidrule(lr){2-6} \cmidrule(lr){7-11} \cmidrule(lr){12-16}
\multirow{2}{*}{Settings/Metrics} & \multicolumn{3}{c}{Distribution}                                                 & \multicolumn{2}{c}{Rank-rank}                                            & \multicolumn{3}{c}{Score}                                                         & \multicolumn{2}{c}{Rank-rank}                                            & \multicolumn{3}{c}{Score}                                                         & \multicolumn{2}{c}{Rank-rank}                                            \\ \cmidrule(lr){2-4} \cmidrule(lr){5-6} \cmidrule(lr){7-9} \cmidrule(lr){10-11} \cmidrule(lr){12-14} \cmidrule(lr){15-16}
                                  & \multicolumn{1}{c}{KL ↓} & \multicolumn{1}{c}{JSD ↓} & \multicolumn{1}{c}{TVD ↓} & \multicolumn{1}{c}{$\tau$ ↑} & \multicolumn{1}{c}{$\rho$ ↑ } & \multicolumn{1}{c}{RMSE ↓} & \multicolumn{1}{c}{MAE ↓} & \multicolumn{1}{c}{R2 ↑} & \multicolumn{1}{c}{$\tau$ ↑} & \multicolumn{1}{c}{$\rho$ ↑ } & \multicolumn{1}{c}{RMSE ↓} & \multicolumn{1}{c}{MAE ↓} & \multicolumn{1}{c}{R2 ↑} & \multicolumn{1}{c}{$\tau$ ↑} & \multicolumn{1}{c}{$\rho$ ↑ } \\ \midrule
baseline                          & 1,0006                   & 0,2644                    & 0,2776                    & 0,4971                              & 0,5119                             & 0,8630                     & 0,7461                    & 0,1300                   & 0,5451                              & 0,6069                             & 0,9101                     & 0,7417                    & 0,4255                   & 0,5395                              & 0,6283                             \\
\rowcolor{green!20}
HumanEX                           & 0,9408                   & 0,2455                    & 0,2448                    & 0,7411                              & 0,7872                             & 0,8912                     & 0,7730                    & 0,0912                   & 0,4047                              & 0,4377                             & 0,9209                     & 0,7536                    & 0,4205                   & 0,4507                              & 0,5225                             \\ \midrule
R1 - \(\text{CoT}_{\text{parser}}\)                       & 0,9610                   & 0,2576                    & 0,2637                    & 0,5597                              & 0,5966                             & 0,8222                     & 0,7113                    & 0,2429                   & 0,5450                              & 0,6169                             & 0,8849                     & 0,7298                    & 0,4428                   & 0,5716                              & 0,6419                             \\
R1 - \(\text{EL}\)                                & 0,9583                   & 0,2566                    & 0,2625                    & 0,5693                              & 0,6089                             & 0,8164                     & 0,7184                    & 0,2479                   & 0,5611                              & 0,6179                             & 0,8845                     & 0,7298                    & 0,4554                   & 0,5957                              & 0,6492                             \\
R1 - \(\text{EL}_{\text{filter}}\text{-sup}\)                         & \textbf{0,9534}          & \textbf{0,2552}           & \textbf{0,2604}           & \textbf{0,6050}                     & \textbf{0,6408}                    & \textbf{0,7698}            & \textbf{0,6660}           & \textbf{0,3176}          & \textbf{0,6500}                     & \textbf{0,6951}                    & \textbf{0,8646}            & \textbf{0,6956}           & \textbf{0,4937}          & \textbf{0,6114}                     & \textbf{0,6790}                    \\ \midrule
QwQ - \(\text{CoT}_{\text{parser}}\)                            & 0,9504                   & 0,2534                    & 0,2589                    & 0,5698                              & 0,6201                             & 0,8607                     & 0,7248                    & 0,2536                   & 0,6002                              & 0,6346                             & 0,9006                     & 0,7326                    & 0,4329                   & 0,6253                              & 0,6734                             \\
QwQ - \(\text{EL}\)                            & 0,9488                   & 0,2535                    & 0,2583                    & 0,5962                              & 0,6357                             & 0,8597                     & 0,7220                    & 0,2670                   & 0,6089                              & 0,6443                             & 0,8882                     & 0,7317                    & 0,4357                   & 0,6270                              & 0,6966                             \\
QwQ - \(\text{EL}_{\text{filter}}\text{-sup}\)                         & \textbf{0,9471}          & \textbf{0,2528}           & \textbf{0,2552}           & \textbf{0,6104}                     & \textbf{0,6475}                    & \textbf{0,7709}            & \textbf{0,6672}           & \textbf{0,3212}          & \textbf{0,6394}                     & \textbf{0,6830}                    & \textbf{0,8787}            & \textbf{0,7197}           & \textbf{0,4541}          & \textbf{0,6378}                     & \textbf{0,7109}     \\
\bottomrule
\end{tabular}}
\caption{Results for the structure ablation study (QwQ as judge).}
\label{tab:structure_results}
\end{table*}
\section{Discussion and Future Works}
\label{sec:discussion}

\paragraph{Generalizability to Other Open-ended Tasks.}
Since HLV evaluation is already subtle and challenging, we chose to first establish a rigorous and interpretable evaluation framework in closed-form settings. We chose NLI and multiple-choice QA tasks primarily because they have closed output spaces, which makes EL pair extraction and evaluation more tractable and reliable. MCQA formats are also widely used in current LLM evaluation benchmarks \cite{DBLP:conf/iclr/HendrycksBBZMSS21,DBLP:journals/corr/abs-1803-05457,DBLP:conf/nips/WangPNSMHLB19,DBLP:journals/tmlr/SrivastavaRRSAF23}, making our approach broadly applicable in practical settings. In contrast, open-ended tasks—such as summarization or free-form sentiment classification—while very interesting, pose greater challenges for automated evaluation, especially under HLV settings, where subjectivity and ambiguity are pronounced. These tasks typically require extensive human validation, and current metrics for open-ended outputs remain underdeveloped for nuanced human disagreement modeling.

We believe that once rank-based evaluation and EL-based modeling are better understood and validated in these contexts, they can be adapted or extended to open-ended tasks.

\paragraph{Qualitative Comparison Between Human and Machine-generated Explanations.}
Qualitative comparison between LLM- and human-generated explanations is indeed valuable, but also challenging in the HLV setting. Many traditional explanation metrics—such as lexical overlap, syntactic similarity, or semantic similarity \cite{giulianelli-etal-2023-comes}—fail to capture the subtle, multidimensional reasoning signals that impact label variation. As noted in \citet{chen-etal-2025-rose}, even minor lexical choices in explanations can shift annotators' perceived plausibility of labels, and human vs. model explanations often diverge subtly in wording, though these differences may have minimal effect on downstream judgments by LLM judges.

Due to the limitations of existing automatic explanation evaluation metrics, we rely on LLM-as-judge setups to assess the quality of the explanations more comprehensively.

\section{Conclusion}

We have demonstrated that CoTs offer a rich and underexplored source of explanation for modeling human label variation, shifting from the traditional reverse explanation paradigm to the forward, rationale-grounded paradigm. Our proposed pipeline is able to extract high-quality explanation-label pairs by leveraging LLMs and refining them through linguistically-grounded discourse segmentation models. Our results show that combining LLMs with discourse segmenters improves the alignment of model explanations with the inherently diverse perspectives of human annotators. Furthermore, our proposed rank-based evaluation framework reflects a more faithful match to human annotation behavior, moving beyond distributional comparisons.

We believe our findings lay the groundwork for more robust, explanation-driven, and linguistically-enhanced approaches to understanding and evaluating human label variation. While this work only leveraged discourse segmentation, explicitly incorporating discourse relations—such as \textit{contrast} or \textit{causal}—may help and provide deeper insights into how reasoning structures map onto human disagreement, ambiguity, and aid interpretation.


\section*{Limitations}

One limitation of our approach lies in the use of discourse segmenters that were trained on existing discourse datasets, which may differ in style and content from the CoT reasoning text we analyze. As a result, the segmenter outputs may not optimally reflect the discourse structure inherent to CoTs, which often contain informal, fragmented, or model-specific reasoning styles. Moreover, we did not conduct a comprehensive evaluation of segmenter performance on CoT data but instead relied directly on the segmenter outputs. While the performance of the discourse segmenters is relatively good for English (as shown in Appendix \ref{app:discodisco-training-results}), future work might benefit from developing or fine-tuning these discourse models specifically on the annotated CoT data, which could potentially improve the precision and interpretability of discourse-informed explanation extraction by incorporating discourse relations—especially those aligned with human rhetorical patterns.

\section*{Acknowledgements}
We thank the members of the MaiNLP lab for their insightful feedback on earlier drafts of this paper. 
We specifically appreciate the suggestions of Philipp Mondorf, Soh-Eun Shim, Yupei Du, and Raoyuan Zhao. 
We extend our gratitude to Pingjun Hong for annotating the human validation experiment.
BC acknowledges his membership in the European Laboratory for Learning and Intelligent Systems (ELLIS) PhD program.
BP and YJL\footnote{Work was carried out while at MaiNLP, LMU Munich.} are supported by ERC Consolidator Grant DIALECT 101043235.
AK is supported by the UK Research and Innovation (UKRI) Frontier Research Grant EP/Y031350/1 EQUATE (the UK government's funding guarantee for ERC Advanced Grants).


\appendix
\section{Training Details and Performance of DisCoDisCo}
\label{app:discodisco-training-results}

We train DisCoDisCo \cite{gessler-etal-2021-discodisco}, the winning system of the DISRPT 2021 Shared Task \cite{zeldes-etal-2021-disrpt} using the latest DISRPT 2023 Shared Task data \cite{braud-etal-2023-disrpt}. Specifically, for the discourse unit segmentation model, we use the English GUM corpus \cite{Zeldes2017} which contains multiple genres, which has been proved to achieve better model generalizability when trained on genre-diverse data for discourse parsing \cite{liu-zeldes-2023-cant}. For the connective detection model, we use the PDTB v3 data in DISRPT, the largest English connective dataset to date. Table \ref{tab:disco-results} shows the performance of both models on their respective test partition averaged over five runs.

\begin{table}[h]
\centering
\resizebox{\columnwidth}{!}{%
\begin{tabular}{@{}c|ccc@{}}
\toprule
   Model                  & Precision & Recall & F1    \\ \midrule
EDU segmentation     & 84.06     & 80.66  & 82.32 \\
connective detection & 94.20     & 95.26  & 94.73 \\ \bottomrule
\end{tabular}%
}
\caption{Performance of the Two Discourse Segmenters.}
\label{tab:disco-results}
\end{table}

\section{Detailed Implementation of the Proposed {\pipename} Pipeline}
\label{app:pipeline-prompt}

This section describes the implementation details of our proposed {\pipename} pipeline. As we consider the CoT process to be a forward reasoning procedure aligned with human annotation, we construct a task-specific prompt for each of the three tasks—VariErr NLI, SIQA, and CQA—that adheres to the forward human annotation process. These prompts are shown in Table~\ref{tab:app1-task-prompt}. We additionally provide the corresponding prompt used for the direct explanation generation method (GenEX) following \citet{chen-etal-2025-rose}. Base LLMs and reasoning-tuned LLMs both use the GenEX prompt respectively to obtain the corresponding Base-GenEX and Reasoning-GenEX settings.

By combining the prompt in Table~\ref{tab:app1-task-prompt} with the input instance (i.e., question and candidate options), we query the reasoning-tuned LLM to generate a CoT reasoning trace, as expressed in Equation~\ref{eq:cot_generation}. Subsequently, we further prompt the reasoning-tuned LLM to parse the generated CoT into supporting and opposing statements.\footnote{Preliminary experiments suggest that when only asked to extract supporting statements, the reasoning-tuned LLM tends to mix in some opposing content. By explicitly prompting the model to output supporting and opposing statements separately, we significantly reduce this ambiguity.} 
The upper portion of Table~\ref{tab:app1-llm-parser-prompt} presents the specific parsing prompt. This step utilizes the reasoning-tuned LLM itself to parse its prior output and generate the parsed CoT, denoted as CoT$_{\text{parser}}$.

Due to the diversity and randomness inherent in LLM outputs, the format of CoT$_{\text{parser}}$ is highly variable and difficult to post-process. Therefore, we leverage the \texttt{JSON} output capabilities of the base LLM associated with the reasoning-tuned LLM. Specifically, we include a system prompt instructing the base LLM to produce a well-structured \texttt{JSON} output adhering to a predefined format (prompt at the bottom of Table~\ref{tab:app1-llm-parser-prompt}. This allows for easier downstream processing into the EL pairs as shown in Equation~\ref{eq:structured_el}.\!\footnote{Even after obtaining the \texttt{JSON} outputs, we further perform post-processing to ensure that the options correctly align with their respective labels.}

\begin{table*}[t]
\centering
\resizebox{\textwidth}{!}{
\begin{tabular}{P{0.15\textwidth}|P{0.85\textwidth}}
\toprule 
\textbf{Datasets} & \textbf{Prompts} \\
\midrule
VariErr NLI \newline CoT & Please determine whether the following statement is true (entailment), undetermined (neutral), or false (contradiction) given the context below and select ONE of the listed options and start your answer with a single letter. \newline Context: \{premise\} \newline Statement: \{hypothesis\} \newline A. Entailment \newline B. Neutral \newline C. Contradiction \newline Answer: \\ \midrule
VariErr NLI \newline GenEX &  You are an expert in Natural Language Inference (NLI). Please list all possible explanations why the following statement is 
\{target-label\} given the context below without introductory phrases. \newline Context: \{premise\} \newline Statement: \{hypothesis\} \newline Answer: \\ \midrule
SIQA \newline CoT & Please read the following social scenario and the accompanying question, choose the most appropriate answer from the options provided and start your answer with a single letter. \newline Scenario: \{scenario\} \newline Question: \{question\} \newline A. \{answerA\} \newline B. \{answerB\} \newline C. 
\{answerC\} \newline Answer: \\ \midrule
SIQA \newline GenEX & You are an expert in social intelligence question answering. Please list all possible explanations why the most appropriate answer is 
\{target-label\} given the following social scenario and the accompanying question below without introductory phrases.  \newline Scenario: \{scenario\} \newline Question: \{question\} \newline Answer: \\ \midrule
CQA \newline CoT & Please read the following question, choose the most appropriate answer from the options provided and start your answer with a single letter. \newline Question: \{question\} \newline A. \{answerA\} \newline B. \{answerB\} \newline C. \{answerC\} \newline D. \{answerD\} \newline E. \{answerE\} \newline Answer: \\ \midrule
CQA \newline GenEX & You are an expert in commonsense question answering. Please list all possible explanations why the most appropriate answer is \{target-label\} given the question below without introductory phrases. \newline Question: 
\{question\} \newline Answer: \\
\bottomrule
\end{tabular}}
\caption{
The forward task-specific prompts for CoT or direct explanation generation.
}
\label{tab:app1-task-prompt}
\end{table*}

\begin{table*}[t]
\centering
\resizebox{\textwidth}{!}{
\begin{tabular}{P{0.15\textwidth}|P{0.85\textwidth}}
\toprule 
\textbf{Explanations} & \textbf{Prompts} \\
\midrule
CoT$_{\text{parser}}$ & The content of your reasoning process is below: \newline \{CoT\} \newline Please extract and list all the original sentences from the aforementioned reasoning process that support and oppose each option separately. \\ \midrule
EL &     \textbf{system prompt:} \newline
    Convert the given markdown into a structured JSON where each option has two keys: support and oppose. Each key should map to a list of statements from the markdown that either support or oppose that option. \newline
    
    EXAMPLE JSON OUTPUT: \newline
    \{ \newline
      "Option A": \{ \newline
        "support": ["SentenceA.1","SentenceA.2"], \newline
        "oppose": ["SentenceA.3"] \newline
      \}, \newline
        "Option B": \{ \newline
        "support": ["SentenceB.1"], \newline
        "oppose": [] \newline
      \}, \newline
      ... \newline
    \}
     \\ 
\bottomrule
\end{tabular}}
\caption{
Prompts for LLM parser and \texttt{JSON} structuring.
}
\label{tab:app1-llm-parser-prompt}
\end{table*}
\section{Detailed Metrics and Results for Human Validation}
\label{app:human-anno-eval-dimensions}

\subsection{Metrics Calculation}
\label{appsub:metrics_human}
To evaluate the similarity between textual explanations, we follow prior work~\cite{giulianelli-etal-2023-comes} and adopt three metrics that capture different linguistic aspects: {Lexical}, {Syntactic}, and {Semantic} similarities. In addition, we extend this framework by incorporating the {Levenshtein Ratio} as a fourth metric. All metrics are implemented as distance functions normalized to the range $[0,1]$, where higher values indicate greater dissimilarity. Their definitions and computation methods are detailed below.

\paragraph{Lexical Similarity.}
Lexical similarity is defined based on the overlap of $n$-grams between two strings. For $n \in \{1,2,3\}$, we compute the sets of $n$-grams for each string and measure the proportion of shared $n$-grams:

\begin{equation}
    S_{\text{lexical}} = \frac{|G_n(A) \cap G_n(B)|}{|G_n(A) \cup G_n(B)|},
\end{equation}

\noindent where $G_n(X)$ denotes the set of $n$-grams extracted from string $X$. This metric rewards surface-level lexical overlap.

\paragraph{Syntactic Similarity.}
Syntactic similarity follows the same formulation as lexical similarity but operates on sequences of part-of-speech (POS) tags rather than surface tokens. POS tagging is performed using the \texttt{spaCy} pipeline:\footnote{From spaCy, \texttt{en\_core\_web\_md}~\cite{honnibal2020spacy}.}

\begin{equation}
    S_{\text{syntactic}} = \frac{|T_n(A) \cap T_n(B)|}{|T_n(A) \cup T_n(B)|},
\end{equation}

\noindent where $T_n(X)$ denotes the set of POS tag $n$-grams of sentence $X$.

\paragraph{Semantic Similarity.}
Semantic similarity is computed using cosine similarity between sentence embeddings. We use the model following~\citet{reimers-gurevych-2019-sentence} to obtain dense vector representations $v_A$ and $v_B$:\footnote{\texttt{sentence-transformers/all-distilroberta-v1.}}

\begin{equation}
    S_{\text{semantic}} = \frac{v_A \cdot v_B}{\|v_A\| \|v_B\|},
\end{equation}

Cosine similarity returns values in $[-1, 1]$, but since embeddings from this model are non-negative, it typically yields values in $[0, 1]$.

\paragraph{Levenshtein Ratio.}
We also include a character-level similarity measure: the Levenshtein Ratio. Let $\text{lev}(A, B)$ denote the Levenshtein distance, i.e., the minimum number of character-level edits (insertions, deletions, substitutions) needed to transform string $A$ into $B$. The Levenshtein Ratio is defined as:

\begin{equation}
    S_{\text{lev}} = 1 - \frac{\text{lev}(A, B)}{\max(|A|, |B|)},
\end{equation}

\noindent where $|A|$ and $|B|$ are the lengths of the strings. This score approaches 1 when the strings are nearly identical and decreases as they diverge.

All similarity scores are bounded in $[0,1]$ and are designed such that higher scores indicate stronger similarity. This unified setup supports a nuanced, multi-level analysis of explanation similarity and invites future extensions involving additional linguistic or pragmatic metrics.

\subsection{Detailed Scores for Human Validation}

We here introduce the detailed procedure for computing the human validation scores. Assume that within the Explanation-Label (EL) pairs, there are \(k\) distinct labels. For each label, there exist two types of explanation sets: \textit{support} and \textit{oppose}. The same structure holds for the human-annotated explanation-label pairs, denoted as \(\text{EL}_{\text{human}}\).

For each label \(l \in \{1, \dots, k\}\), and for each stance \(s \in \{\text{support}, \text{oppose}\}\), we compare the corresponding explanation sets from EL and \(\text{EL}_{\text{human}}\). Let these be denoted as:
\begin{equation}
\text{EXSet}_{\text{EL}}^{(l,s)} \quad \text{and} \quad \text{EXSet}_{\text{EL}_{\text{human}}}^{(l,s)},
\end{equation}

The similarity score for each such pair is computed as follows:

\begin{itemize}
    \item If one of the sets is empty while the other is non-empty, assign a score of 0.
    \item If both sets are empty, assign a score of 1.
    \item If both sets are non-empty:
    
$\mathrm{i}$) For each explanation \(e \in \text{EXSet}_{\text{EL}}^{(l,s)}\), compute its similarity with all explanations \(h \in \text{EXSet}_{\text{EL}_{\text{human}}}^{(l,s)}\) using the four metrics described in \S~\ref{appsub:metrics_human}.
 
$\mathrm{ii}$) For each explanation \(e\), define its score as the maximum of its average similarity across metrics:
        \begin{equation}
        \text{sim}(e) = \max_{h \in \text{EXSet}_{\text{EL}_{\text{human}}}^{(l,s)}} \text{avg\_sim}(e, h),
        \end{equation}
        where \(\text{avg\_sim}(e, h)\) denotes the mean of the four similarity metrics.
        
$\mathrm{iii}$) The final similarity score for the pair \((l, s)\) is the average of \(\text{sim}(e)\) over all \(e \in \text{EXSet}_{\text{EL}}^{(l,s)}\):
        \begin{equation}
        \text{Score}^{(l,s)} = \frac{1}{|\text{EXSet}_{\text{EL}}^{(l,s)}|} \sum_{e \in \text{EXSet}_{\text{EL}}^{(l,s)}} \text{sim}(e),
        \end{equation}
\end{itemize}

After calculating the scores for all \(2k\) explanation set pairs (i.e., each label's support and oppose explanations), we compute the average to obtain the similarity score between EL and \(\text{EL}_{\text{human}}\) for a single instance:

{\small
\begin{equation}
\text{S\_instance} = \frac{1}{2k} \sum_{l=1}^k \left( \text{Score}^{(l, \text{support})} + \text{Score}^{(l, \text{oppose})} \right).
\end{equation}
}

Finally, we average the instance-level scores over all instances in the dataset to obtain the overall similarity score. Importantly, although \(\text{avg\_sim}\) is used only for selecting the best match per explanation, the scores for each of the four individual metrics are also recorded and averaged across all explanations and instances. The final results for each of the four metrics are reported in Table~\ref{tab:validation-1}.

\begin{table*}[t]
\centering
\resizebox{\linewidth}{!}{
\begin{tabular}{lccc|ccc|cc|c|cc}
\toprule
\multicolumn{1}{c}{\multirow{2}{*}{\textbf{LLMs - Datasets}}} & \multicolumn{3}{c|}{\textbf{Lexical}} & \multicolumn{3}{c|}{\textbf{Syntactic}} & \multicolumn{2}{c|}{\textbf{Semantic}} &  \textbf{Levenshtein Ratio} & \multicolumn{2}{|c}{\textbf{AVG}}\\ \cmidrule(lr){2-12} 
\multicolumn{1}{c}{}          & {n = 1$\uparrow$} & {n = 2 $\uparrow$} & {n = 3$\uparrow$}                                     & {n = 1$\uparrow$ } & {n = 2$\uparrow$ } & {n = 3$\uparrow$ }         & {Cos.$\uparrow$ }  &  {Euc.$\uparrow$ }  & { ratio $\uparrow$}  & {equal-avg $\uparrow$} & {weight-avg $\uparrow$}  \\
\midrule
\multicolumn{12}{l}{{\textbf{DeepSeek R1 - VariErr NLI - CoT}}}   \\
\multicolumn{12}{l}{\emph{all}}   \\
\, \(\text{EL}\) &  0,6877 & 0,6249 & 0,5982 & 0,8209 & 0,7045 & 0,6468 & 0,7202 & 0,6877 & 0,6470 & 0,6820 & 0,6780 \\
\, \(\text{EL}_{\text{filter}}\) & 0,8309 & 0,7883 & 0,7756 & 0,9119 & 0,8295 & 0,7831 & 0,8265 & 0,7551 & 0,7943 & 0,8106 & 0,8062 \\
\multicolumn{12}{l}{\emph{only-support}}   \\  
\, \(\text{EL-sup}\) &  0,7152 & 0,6607 & 0,6432 & 0,8233 & 0,7141 & 0,6598 & 0,7192 & 0,6779 & 0,6793 & 0,6992 & 0,6958  \\
\, \(\text{EL}_{\text{filter}}\text{-sup}\)  & 0,8514 & 0,8108 & 0,7995 & 0,9199 & 0,8477 & 0,8060 & 0,8410 & 0,7668 & 0,8232 & 0,8296 & 0,8264\\
\multicolumn{12}{l}{{\textbf{DeepSeek R1 - SIQA - CoT}}}   \\
\multicolumn{12}{l}{\emph{all}}  \\
\, \(\text{EL}\) &  0,8095 & 0,7632 & 0,7471 & 0,8920 & 0,8071 & 0,7720 & 0,8228 & 0,7575 & 0,7364 & 0,7897 & 0,7809  \\
\, \(\text{EL}_{\text{filter}}\)  & 0,8947 & 0,8823 & 0,8782 & 0,9197 & 0,8863 & 0,8749 & 0,8866 & 0,7913 & 0,8712 & 0,8761 & 0,8722\\
\multicolumn{12}{l}{\emph{only-support}}   \\  
\, \(\text{EL-sup}\) &  0,8360 & 0,7947 & 0,7856 & 0,9046 & 0,8257 & 0,7921 & 0,8424 & 0,7611 & 0,8081 & 0,8167 & 0,8140  \\
\, \(\text{EL}_{\text{filter}}\text{-sup}\)  & 0,9000 & 0,8895 & 0,8861 & 0,9220 & 0,8923 & 0,8821 & 0,8938 & 0,7955 & 0,8810 & 0,8825 & 0,8791 \\
\multicolumn{12}{l}{{\textbf{DeepSeek R1 - CQA - CoT}}}   \\
\multicolumn{12}{l}{\emph{all}}   \\
\, \(\text{EL}\) & 0,8400 & 0,7988 & 0,7843 & 0,9067 & 0,8399 & 0,8071 & 0,8408 & 0,7856 & 0,7771 & 0,8200 & 0,8123  \\
\, \(\text{EL}_{\text{filter}}\)  & 0,8887 & 0,8749 & 0,8713 & 0,9190 & 0,8907 & 0,8721 & 0,8676 & 0,7722 & 0,8591 & 0,8684 & 0,8628\\
\multicolumn{12}{l}{\emph{only-support}}   \\  
\, \(\text{EL-sup}\) &  0,8536 & 0,8295 & 0,8250 & 0,8860 & 0,8515 & 0,8342 & 0,8585 & 0,8137 & 0,8356 & 0,8431 & 0,8412  \\
\, \(\text{EL}_{\text{filter}}\text{-sup}\)  & 0,8962 & 0,8828 & 0,8797 & 0,9210 & 0,8967 & 0,8796 & 0,8724 & 0,7765 & 0,8692 & 0,8749 & 0,8697 \\
\bottomrule
\end{tabular}}
\caption{Results for the validation based on human annotated subsets.}\label{tab:validation-1}
\end{table*}

\section{Details of Ranking Generation Methods}
\label{app:ranking-method}

Here we elaborate on the implementation details of the three LLM-judge-based ranking generation methods introduced in Section~\ref{subsec:ranking-gen-methods}. Note that for all methods, the final ranking is obtained by averaging the rankings from three independent runs.

\paragraph{Direct Ranking.} In this method, we prompt the LLM to directly generate a ranking. The prompts used for different tasks are listed in Table~\ref{tab:app4-direct_prompt}. After receiving a space-separated list of options, we process the output as follows: if indices for all options are present, we rank them according to the order in which they appear. If only a subset of indices is provided, the missing options are assigned the lowest possible rank (i.e., tied for last place).

\paragraph{First-Token-Logits Ranking.} The prompt used in this method is identical to the one used for forward chain-of-thought generation (see Table~\ref{tab:app1-task-prompt}). However, in this case, we focus on the first token of the LLM's answer. Following the method proposed in \citet{chen-etal-2025-rose,chen-etal-2024-seeing}, we extract the scores corresponding to each option index from the first-token logits. We then normalize these scores to obtain a probability distribution over the labels. This distribution can be used for distribution-based similarity evaluation or converted into rankings.

\paragraph{Scoring-Based Ranking.} In this approach, we ask the LLM judges to assign a likelihood score from 1 to 5 for each option, with higher scores indicating higher plausibility. The prompt used for this setting is shown in Table~\ref{tab:app4-score_prompt}. These scores can be used for score-based similarity evaluation or transformed into rankings for ranking-based evaluation.

\paragraph{}
To evaluate the performance of the explanations, we augment all the above prompts with explanation content and instruct the LLM judges to take these rationales into account when making their decisions. We take the EL injection prompt for VariErr NLI as an example in Table~\ref{tab:app4-EL-injection_prompt}. Both SIQA and CQA adopt similar prompts. 

All the prompts described above can be readily generalized to a wide range of tasks, such as summarization, sentiment analysis, cultural evaluation and even visual question answering \cite{DBLP:journals/jair/UmaFHPPP21,DBLP:journals/corr/abs-2505-21693,DBLP:conf/aaai/LanFP25, DBLP:journals/corr/abs-2505-23798}. In addition, insights from linguistics may further refine and improve these prompts. For example, as discussed in \citet{jiang-etal-2023-ecologically,DBLP:journals/corr/abs-2505-22848}, adopting a more fine-grained taxonomy in the NLI task could decompose the currently broad categories of entailment, neutral, and contradiction, thereby yielding a more precise label distribution.

\begin{table*}[t]
\centering
\resizebox{\textwidth}{!}{
\begin{tabular}{P{0.15\textwidth}|P{0.85\textwidth}}
\toprule 
\textbf{Datasets} & \textbf{Prompts} \\
\midrule
VariErr NLI  & 
Please assess whether the following statement is true (entailment), undetermined (neutral), or false (contradiction) given the context below, rank all the following options from most appropriate to least appropriate. Only output the letters representing the options, separated by spaces. \newline Context: \{premise\} \newline Statement: \{hypothesis\} \newline A. Entailment \newline B. Neutral \newline C. Contradiction \newline Answer: 
\\ \midrule
SIQA & 
Please read the following social scenario and the accompanying question, rank all the following options from best to worst based on relevance and appropriateness. Only output the letters representing the options, separated by spaces. \newline Scenario: \{scenario\} \newline Question: \{question\} \newline A. \{answerA\} \newline B. \{answerB\} \newline C. 
\{answerC\} \newline Answer:
\\ \midrule
CQA & 
Please read the following question, rank all the following options from best to worst based on relevance and appropriateness. Only output the letters representing the options, separated by spaces. \newline Question: \{question\} \newline A. \{answerA\} \newline B. \{answerB\} \newline C. \{answerC\} \newline D. \{answerD\} \newline E. \{answerE\} \newline Answer: 
\\
\bottomrule
\end{tabular}}
\caption{
The prompts for the direct ranking method across three datasets.
}
\label{tab:app4-direct_prompt}
\end{table*}

\begin{table*}[t]
\centering
\resizebox{\textwidth}{!}{
\begin{tabular}{P{0.15\textwidth}|P{0.85\textwidth}}
\toprule 
\textbf{Datasets} & \textbf{Prompts} \\
\midrule
VariErr NLI  & 
Please rate the following answer based on its plausibility in representing the relationship between the context and the statement on the 5-Point Scale rating as below. Only output a single integer corresponding to your evaluation. \newline Context: \{premise\} \newline Statement: \{hypothesis\} \newline Answer: \{target-label\} \newline Plausibility Ratings: \newline 1 = Impossible \newline 2 = Technically Possible \newline 3 = Plausible \newline 4 = Likely \newline 5 = Very Likely \newline Rating:
\\ \midrule
SIQA & 
Please read the following social scenario and the accompanying question, rate the plausibility of the answer on the 5-Point Scale rating as below. Only output a single integer corresponding to your evaluation. \newline Scenario: \{scenario\} \newline Question: \{question\} \newline Answer: \{target-label\} \newline Plausibility Ratings: \newline 1 = Impossible \newline 2 = Technically Possible \newline 3 = Plausible \newline 4 = Likely \newline 5 = Very Likely \newline Rating:
\\ \midrule
CQA & 
Please read the following question, rate the plausibility of the answer on the 5-Point Scale rating as below. Only output a single integer corresponding to your evaluation. \newline Question: \{question\} \newline Answer: \{target-label\} \newline Plausibility Ratings: \newline 1 = Impossible \newline 2 = Technically Possible \newline 3 = Plausible \newline 4 = Likely \newline 5 = Very Likely \newline Rating:
\\
\bottomrule
\end{tabular}}
\caption{
The prompts for the score-based ranking method across three datasets.
}
\label{tab:app4-score_prompt}
\end{table*}

\begin{table*}[t]
\centering
\resizebox{\textwidth}{!}{
\begin{tabular}{P{0.15\textwidth}|P{0.85\textwidth}}
\toprule 
\textbf{Datasets} & \textbf{Prompts} \\
\midrule
Direct Ranking with ELs  & 
Please assess whether the following statement is true (entailment), undetermined (neutral), or false (contradiction) given the context below. Consider relevant perspectives, possible explanations, or reasoning patterns in the following explanations. Rank all the following options from most appropriate to least appropriate. Only output the letters representing the options, separated by spaces. \newline Context: \{premise\} \newline Statement: \{hypothesis\} \newline A. Entailment \newline B. Neutral \newline C. Contradiction \newline Explanations: \{explanation-label-pairs\} \newline Answer: 
\\ \midrule
First-Token-Logits Ranking with ELs & 
Please determine whether the following statement is true (entailment), undetermined (neutral), or false (contradiction) given the context below. Consider relevant perspectives, possible explanations, or reasoning patterns in the following explanations. Select ONE of the listed options and start your answer with a single letter. \newline Context: \{premise\} \newline Statement: \{hypothesis\} \newline A. Entailment \newline B. Neutral \newline C. Contradiction \newline Explanations: \{explanation-label-pairs\} \newline Answer: \\ 
\\ \midrule
Score-Based Ranking with ELs & 
Please rate the following answer based on its plausibility in representing the relationship between the context and the statement on the 5-Point Scale rating as below. Consider relevant perspectives, possible explanations, or reasoning patterns in the following explanations. Only output a single integer corresponding to your evaluation. \newline Context: \{premise\} \newline Statement: \{hypothesis\} \newline Answer: \{target-label\} \newline Plausibility Ratings: \newline 1 = Impossible \newline 2 = Technically Possible \newline 3 = Plausible \newline 4 = Likely \newline 5 = Very Likely \newline Explanations: \{explanation-label-pairs\} \newline Rating:
\\
\bottomrule
\end{tabular}}
\caption{
The example prompts of the EL injection for VariErr NLI.
}
\label{tab:app4-EL-injection_prompt}
\end{table*}
\section{Details of the Metrics in HLV Evaluation}
\label{app:HLV-metric}

This section provides a detailed explanation of the calculation formulas for all the metrics introduced in \S\ref{subsec:all_metrics_for_hlv}.

\subsection{Rank Correlation Metrics}

Let \( (x_i, y_i) \) for \( i = 1, \dots, n \) be paired ranks from two sources (e.g., human vs. model).

\paragraph{Kendall’s \(\tau\)} \cite{kendall1938new} Measures the difference between the number of concordant and discordant pairs:

\begin{equation}
\tau = \frac{C - D}{\frac{1}{2}n(n-1)} ,
\end{equation}

\noindent where \( C \) is the number of concordant pairs and \( D \) is the number of discordant pairs.

\paragraph{Spearman’s \(\rho\)} \cite{spearman1961proof} Measures the Pearson correlation between rank variables:

\begin{equation}
\rho = 1 - \frac{6 \sum_{i=1}^{n} d_i^2}{n(n^2 - 1)} ,
\end{equation}

\noindent where \( d_i = x_i - y_i \) is the difference between the ranks.

\subsection{Distribution-Based Metrics}

For probability distributions (from VariErr NLI), we use:

\begin{itemize}
    \item \textbf{Kullback-Leibler Divergence (KL)}~\cite{kullback1951information}
    \item \textbf{Jensen-Shannon Distance (JSD)}~\cite{DBLP:journals/tit/EndresS03}
    \item \textbf{Total Variation Distance (TVD)}~\cite{DBLP:books/daglib/0018090}
\end{itemize}

Given discrete distributions \(P\) and \(Q\):

\begin{equation}
D_{\text{KL}}(P \| Q) = \sum_{x \in \mathcal{X}} P(x) \log \frac{P(x)}{Q(x)} ,
\end{equation}

{\small
\begin{equation}
D_{\text{JSD}}(P \| Q) = \sqrt{\frac{1}{2} \left( D_{\text{KL}}(P \| M) + D_{\text{KL}}(Q \| M) \right)} ,
\end{equation}
}

\noindent where \( M = \frac{1}{2}(P + Q) \).

\begin{equation}
D_{\text{TVD}}(P, Q) = \frac{1}{2} \sum_{x \in \mathcal{X}} |P(x) - Q(x)| ,
\end{equation}

\subsection{Scalar-Based Metrics}

For scalar scores (e.g., from SIQA and CQA), we use:

\begin{itemize}
    \item \textbf{Root Mean Squared Error (RMSE)}~\cite{hyndman2006another}
    \item \textbf{Mean Absolute Error (MAE)}~\cite{willmott2005advantages}
    \item \textbf{Coefficient of Determination (\(R^2\))}~\cite{steel1960principles}
\end{itemize}

\begin{equation}
\text{RMSE} = \sqrt{\frac{1}{n} \sum_{i=1}^{n} (y_i - \hat{y}_i)^2} ,
\end{equation}

\begin{equation}
\text{MAE} = \frac{1}{n} \sum_{i=1}^{n} |y_i - \hat{y}_i| ,
\end{equation}

\begin{equation}
R^2 = 1 - \frac{\sum_{i=1}^{n} (y_i - \hat{y}_i)^2}{\sum_{i=1}^{n} (y_i - \bar{y})^2} .
\end{equation}

\noindent where \( y_i \) is the human annotation, \( \hat{y}_i \) is the model prediction, and \( \bar{y} \) is the mean of human annotations.

\section{HLV Evaluation Full Results}
\label{app:all_results}

In this section, we report the full HLV evaluation results across all settings and datasets. All the result figures and tables presented in \S\ref{sec:results} are derived from the detailed scores provided here. Specifically, the results for VariErr NLI are presented in Table~\ref{tab:applast-HLV-varierr}, SIQA in Table~\ref{tab:applast-HLV-siqa}, and CQA in Table~\ref{tab:applast-HLV-cqa}. All rankings, scores, and distributions from LLM judges are averaged over three independent runs. For VariErr NLI, the gold human distributions and rankings are computed as the average across annotations from MNLI, VariErr NLI, and Chaos NLI, as described in \S\ref{sec:datasets}. For SIQA and CQA, the gold human label scores are obtained by averaging the scores provided by five annotators for each corresponding label.

\begin{table*}[t]
\centering
\resizebox{0.65\linewidth}{!}{
\begin{tabular}{lccccccccc}
\toprule
\multicolumn{1}{c}{\multirow{2}{*}{Settings/Metrics}} & \multicolumn{3}{c}{Distribution}                    & \multicolumn{2}{c}{Rank-rank}     & \multicolumn{2}{c}{Rank-logits}   & \multicolumn{2}{c}{Rank-score}    \\ \cmidrule(lr){2-4} \cmidrule(lr){5-6} \cmidrule(lr){7-8} \cmidrule(lr){9-10} 
\multicolumn{1}{c}{}                                  & KL ↓            & JSD ↓           & TVD ↓           & $\tau$ ↑          & $\rho$ ↑  & $\tau$ ↑           & $\rho$ ↑  & $\tau$ ↑          & $\rho$ ↑ \\ \midrule
\multicolumn{10}{l}{{qwen as judge}} \\
baseline                                              & 1,0006          & 0,2644          & 0,2776          & 0,4971          & 0,5119          & 0,4619          & 0,5085          & 0,3190          & 0,3452          \\
HumanEX                                               & 0,9408          & 0,2455          & 0,2448          & 0,7411          & 0,7872          & 0,6574          & 0,7151          & 0,3864          & 0,4151          \\
V3 GenEX                                              & 0,9835          & 0,2626          & 0,2737          & 0,5071          & 0,5334          & 0,4648          & 0,5269          & 0,2817          & 0,2980          \\
\multicolumn{10}{l}{{R1}} \\
GenEX                                                 & 0,9733          & 0,2615          & 0,2716          & 0,5142          & 0,5557          & 0,4688          & 0,5321          & 0,2902          & 0,3078          \\
CoT                                                   & 0,9565          & 0,2590          & 0,2655          & 0,5129          & 0,5421          & 0,4731          & 0,5399          & 0,3933          & 0,4058          \\
\(\text{CoT}_{\text{parser}}\)                                             & 0,9610          & 0,2576          & 0,2637          & 0,5597          & 0,5966          & 0,4786          & 0,5404          & 0,4014          & 0,4187          \\
EL                                                    & 0,9583          & 0,2566          & 0,2625          & 0,5693          & 0,6089          & 0,4928          & 0,5539          & 0,4064          & 0,4365          \\
\(\text{EL}_{\text{filter}}\)                                                 & \textbf{0,9515} & 0,2558          & 0,2611          & 0,5708          & 0,6352          & 0,5289          & 0,5802          & \textbf{0,4388} & 0,4480          \\
EL-sup                                                & 0,9566          & 0,2564          & 0,2621          & 0,5905          & 0,6260          & 0,5037          & 0,5619          & 0,4122          & 0,4377          \\
\(\text{EL}_{\text{filter}}\text{-sup}\)                                            & 0,9534          & \textbf{0,2552} & \textbf{0,2604} & \textbf{0,6050} & \textbf{0,6408} & \textbf{0,5604} & \textbf{0,6099} & 0,4213          & \textbf{0,4519} \\
EL-opp                                                & 0,9756          & 0,2590          & 0,2675          & 0,4768          & 0,5071          & 0,4734          & 0,5117          & 0,3658          & 0,3903          \\
\(\text{EL}_{\text{filter}}\text{-opp}\)                                            & 0,9716          & 0,2585          & 0,2663          & 0,4898          & 0,5231          & 0,4785          & 0,5171          & 0,3779          & 0,4032          \\
Qwen-Max GenEX                                        & 0,9833          & 0,2617          & 0,2723          & 0,5019          & 0,5459          & 0,4743          & 0,5084          & 0,2807          & 0,3006          \\
\multicolumn{10}{l}{{QwQ}} \\
GenEX                                                 & 0,9620          & 0,2576          & 0,2668          & 0,5701          & 0,6008          & 0,4921          & 0,5253          & 0,2608          & 0,2759          \\
CoT                                                   & 0,9515          & 0,2543          & 0,2606          & 0,5738          & 0,6152          & 0,5095          & 0,5383          & 0,4004          & 0,4232          \\
\(\text{CoT}_{\text{parser}}\)                                             & 0,9504          & 0,2534          & 0,2589          & 0,5698          & 0,6201          & 0,5183          & 0,5491          & 0,4022          & 0,4309          \\
EL                                                    & 0,9488          & 0,2535          & 0,2583          & 0,5962          & 0,6357          & 0,5260          & 0,5534          & 0,4200          & 0,4506          \\
\(\text{EL}_{\text{filter}}\)                                                 & \textbf{0,9409} & \textbf{0,2515} & 0,2567          & 0,6063          & 0,6369          & 0,5580          & \textbf{0,6161} & 0,4675          & 0,5027          \\
EL-sup                                                & 0,9445          & 0,2533          & 0,2582          & 0,6023          & 0,6386          & 0,5286          & 0,5871          & 0,4475          & 0,4771          \\
\(\text{EL}_{\text{filter}}\text{-sup}\)                                            & 0,9471          & 0,2528          & \textbf{0,2552} & \textbf{0,6104} & \textbf{0,6475} & \textbf{0,5637} & 0,6129          & \textbf{0,5287} & \textbf{0,5685} \\
EL-opp                                                & 0,9647          & 0,2572          & 0,2652          & 0,4937          & 0,5269          & 0,4570          & 0,5123          & 0,3741          & 0,3937          \\
\(\text{EL}_{\text{filter}}\text{-opp}\)                                            & 0,9547          & 0,2564          & 0,2639          & 0,4904          & 0,5281          & 0,5034          & 0,5541          & 0,4197          & 0,4095          \\ \midrule
\multicolumn{10}{l}{{llama as judge}} \\
baseline                                              & 1,2415          & 0,2962          & 0,3207          & 0,4067          & 0,4409          & 0,4324          & 0,4739          & 0,0788          & 0,0809          \\
HumanEX                                               & 1,2032          & 0,2883          & 0,3081          & 0,4392          & 0,4640          & 0,5987          & 0,6672          & 0,1591          & 0,1689          \\
V3 GenEX                                              & 1,2561          & 0,2982          & 0,3231          & 0,1716          & 0,1672          & 0,2079          & 0,2205          & 0,0613          & 0,0670          \\
\multicolumn{10}{l}{{R1}} \\
GenEX                                                 & 1,2580          & 0,2982          & 0,3231          & 0,1499          & 0,1529          & 0,1842          & 0,1987          & 0,0739          & 0,0737          \\
CoT                                                   & 1,1953          & 0,2951          & 0,3187          & 0,4140          & 0,4201          & 0,4574          & 0,5308          & 0,1595          & 0,1674          \\
\(\text{CoT}_{\text{parser}}\)                                             & 1,1925          & 0,2904          & 0,3088          & 0,4178          & 0,4264          & 0,4789          & 0,5337          & 0,1603          & 0,1730          \\
EL                                                    & 1,1883          & 0,2877          & 0,3074          & 0,4180          & 0,4339          & 0,4881          & 0,5412          & 0,1614          & 0,1761          \\
\(\text{EL}_{\text{filter}}\)                                                 & 1,1770          & 0,2874          & 0,3049          & 0,4523          & 0,4864          & 0,4980          & 0,5526          & 0,2619          & 0,2855          \\
EL-sup                                                & 1,1722          & 0,2859          & 0,3048          & 0,4292          & 0,4443          & 0,4931          & 0,5464          & 0,1848          & 0,2051          \\
\(\text{EL}_{\text{filter}}\text{-sup}\)                                            & \textbf{1,0831} & \textbf{0,2744} & \textbf{0,2878} & \textbf{0,4645} & \textbf{0,4967} & \textbf{0,5085} & \textbf{0,5568} & \textbf{0,2677} & \textbf{0,2868} \\
EL-opp                                                & 1,2374          & 0,2953          & 0,3185          & 0,4000          & 0,3848          & 0,4248          & 0,4737          & 0,1295          & 0,1402          \\
\(\text{EL}_{\text{filter}}\text{-opp}\)                                            & 1,2339          & 0,2947          & 0,3178          & 0,4095          & 0,3974          & 0,4374          & 0,4872          & 0,1414          & 0,1573          \\
Qwen-Max GenEX                                        & 1,2552          & 0,2970          & 0,3216          & 0,2650          & 0,2779          & 0,3193          & 0,3601          & 0,1238          & 0,1358          \\
\multicolumn{10}{l}{{QwQ}} \\
GenEX                                                 & 1,2665          & 0,2991          & 0,3242          & 0,1746          & 0,1765          & 0,1798          & 0,1884          & 0,0506          & 0,0546          \\
CoT                                                   & 1,1979          & 0,2916          & 0,3140          & 0,4228          & 0,4593          & 0,5040          & 0,5652          & 0,1953          & 0,1374          \\
\(\text{CoT}_{\text{parser}}\)                                             & 1,1991          & 0,2886          & 0,3095          & 0,4587          & 0,4883          & 0,5054          & 0,5696          & 0,2064          & 0,2242          \\
EL                                                    & 1,1812          & 0,2859          & 0,3047          & 0,4696          & 0,4892          & 0,5136          & 0,5714          & 0,2207          & 0,2413          \\
\(\text{EL}_{\text{filter}}\)                                                 & 1,1004          & 0,2755          & 0,2827          & 0,5125          & 0,5036          & 0,5209          & 0,5802          & 0,2902          & 0,3119          \\
EL-sup                                                & 1,1671          & 0,2836          & 0,3013          & 0,5041          & 0,5352          & 0,5194          & 0,5772          & 0,2344          & 0,2515          \\
\(\text{EL}_{\text{filter}}\text{-sup}\)                                            & \textbf{1,0764} & \textbf{0,2708} & \textbf{0,2827} & \textbf{0,5239} & \textbf{0,5573} & \textbf{0,5212} & \textbf{0,5820} & \textbf{0,3128} & \textbf{0,3446} \\
EL-opp                                                & 1,2392          & 0,2954          & 0,3190          & 0,3585          & 0,3824          & 0,4417          & 0,4924          & 0,1127          & 0,1175          \\
\(\text{EL}_{\text{filter}}\text{-opp}\)                                            & 1,2291          & 0,2938          & 0,3165          & 0,3861          & 0,4092          & 0,5035          & 0,5111          & 0,1241          & 0,1247          \\ \midrule
\multicolumn{10}{l}{{mistral as judge}} \\
baseline                                              & 0,6892          & 0,2611          & 0,2949          & 0,4799          & 0,5096          & 0,4053          & 0,4385          & 0,3209          & 0,3444          \\
HumanEX                                               & 0,6228          & 0,2336          & 0,2430          & 0,4994          & 0,5298          & 0,4376          & 0,4747          & 0,4311          & 0,4553          \\
V3 GenEX                                              & 0,7603          & 0,2603          & 0,2841          & 0,3880          & 0,4101          & 0,3572          & 0,3999          & 0,1926          & 0,1999          \\
\multicolumn{10}{l}{{R1}} \\
GenEX                                                 & 0,8239          & 0,2609          & 0,2816          & 0,4211          & 0,4464          & 0,3547          & 0,3900          & 0,1398          & 0,1511          \\
CoT                                                   & 0,6503          & 0,2512          & 0,2756          & 0,4712          & 0,4979          & 0,4213          & 0,4677          & 0,3765          & 0,4073          \\
\(\text{CoT}_{\text{parser}}\)                                             & 0,6471          & 0,2508          & 0,2712          & 0,4853          & 0,5145          & 0,4330          & 0,4705          & 0,3847          & 0,4166          \\
EL                                                    & 0,6405          & 0,2490          & 0,2710          & 0,4860          & 0,5155          & 0,4342          & 0,4742          & 0,3931          & 0,4189          \\
\(\text{EL}_{\text{filter}}\)                                                 & 0,6334          & 0,2479          & 0,2687          & \textbf{0,4959} & 0,5240          & \textbf{0,4466} & 0,4825          & 0,4009          & 0,4259          \\
EL-sup                                                & 0,6384          & 0,2497          & 0,2697          & 0,4880          & 0,5173          & 0,4416          & 0,4779          & 0,4006          & 0,4281          \\
\(\text{EL}_{\text{filter}}\text{-sup}\)                                            & \textbf{0,6331} & \textbf{0,2476} & \textbf{0,2684} & 0,4942          & \textbf{0,5258} & 0,4439          & \textbf{0,4895} & \textbf{0,4035} & \textbf{0,4285} \\
EL-opp                                                & 0,6671          & 0,2619          & 0,2844          & 0,4542          & 0,4775          & 0,4056          & 0,4402          & 0,3510          & 0,3802          \\
\(\text{EL}_{\text{filter}}\text{-opp}\)                                            & 0,6495          & 0,2546          & 0,2860          & 0,4651          & 0,4812          & 0,4189          & 0,4512          & 0,3596          & 0,4031          \\
Qwen-Max GenEX                                        & 0,8876          & 0,2853          & 0,3208          & 0,3435          & 0,3641          & 0,2436          & 0,2652          & 0,2593          & 0,2783          \\
\multicolumn{10}{l}{{QwQ}} \\
GenEX                                                 & 0,8475          & 0,2645          & 0,2889          & 0,3887          & 0,4112          & 0,3390          & 0,3757          & 0,3563          & 0,3917          \\
CoT                                                   & 0,6275          & 0,2580          & 0,2779          & 0,4873          & 0,5019          & 0,4336          & 0,4732          & 0,3993          & 0,4206          \\
\(\text{CoT}_{\text{parser}}\)                                             & 0,6213          & 0,2497          & 0,2649          & 0,4920          & 0,5212          & 0,4433          & 0,4795          & 0,4048          & 0,4230          \\
EL                                                    & 0,6167          & 0,2473          & 0,2639          & 0,4970          & 0,5269          & 0,4436          & 0,4805          & 0,4186          & 0,4293          \\
\(\text{EL}_{\text{filter}}\)                                                 & \textbf{0,5906} & 0,2445          & 0,2616          & 0,5170          & 0,5481          & 0,4588          & \textbf{0,5067} & 0,4287          & 0,4602          \\
EL-sup                                                & 0,6007          & 0,2444          & 0,2626          & 0,5059          & 0,5363          & 0,4529          & 0,4902          & 0,4276          & 0,4557          \\
\(\text{EL}_{\text{filter}}\text{-sup}\)                                            & 0,6003          & \textbf{0,2437} & \textbf{0,2611} & \textbf{0,5429} & \textbf{0,5756} & \textbf{0,4663} & 0,4957          & \textbf{0,4516} & \textbf{0,4812} \\
EL-opp                                                & 0,6612          & 0,2691          & 0,2890          & 0,4562          & 0,4832          & 0,4202          & 0,4477          & 0,3822          & 0,4091          \\
\(\text{EL}_{\text{filter}}\text{-opp}\)                                            & 0,6298          & 0,2681          & 0,2868          & 0,4752          & 0,5037          & 0,4309          & 0,4637          & 0,3833          & 0,4100         \\
\bottomrule
\end{tabular}}
\caption{All HLV evaluation results on VariErr NLI dataset.}
\label{tab:applast-HLV-varierr}
\end{table*}

\begin{table*}[t]
\centering
\resizebox{0.65\linewidth}{!}{
\begin{tabular}{lccccccccc}
\toprule
\multicolumn{1}{c}{\multirow{2}{*}{Settings/Metrics}} & \multicolumn{3}{c}{Score}                    & \multicolumn{2}{c}{Rank-rank}     & \multicolumn{2}{c}{Rank-logits}   & \multicolumn{2}{c}{Rank-score}    \\ \cmidrule(lr){2-4} \cmidrule(lr){5-6} \cmidrule(lr){7-8} \cmidrule(lr){9-10} 
\multicolumn{1}{c}{}                                  & RMSE ↓            & MAE ↓           & $R^2$ ↑           & $\tau$ ↑          & $\rho$ ↑  & $\tau$ ↑           & $\rho$ ↑  & $\tau$ ↑          & $\rho$ ↑ \\ \midrule
\multicolumn{10}{l}{{qwen as judge}} \\
baseline       & 0,8630          & 0,7461          & 0,1300           & 0,5451          & 0,6069          & 0,5500          & 0,6083          & 0,6568          & 0,6924          \\
HumanEX        & 0,8912          & 0,7730          & 0,0912           & 0,4047          & 0,4377          & 0,5258          & 0,5801          & 0,6537          & 0,6904          \\
V3 GenEX       & 1,0422          & 0,9076          & -0,2196          & 0,4708          & 0,5207          & 0,5187          & 0,5647          & 0,5383          & 0,5736          \\
\multicolumn{10}{l}{{R1}} \\
GenEX          & 0,9728          & 0,8473          & -0,0633          & 0,4577          & 0,5148          & 0,5085          & 0,5650          & 0,5668          & 0,5974          \\
CoT            & 0,8759          & 0,7582          & 0,1165           & 0,5453          & 0,6150          & 0,5482          & 0,6171          & 0,6661          & 0,7000          \\
\(\text{CoT}_{\text{parser}}\)      & 0,8222          & 0,7113          & 0,2429           & 0,5450          & 0,6169          & 0,5509          & 0,6212          & 0,6922          & 0,7330          \\
EL             & 0,8164          & 0,7184          & 0,2479           & 0,5611          & 0,6179          & 0,5671          & 0,6292          & 0,6411          & 0,6756          \\
\(\text{EL}_{\text{filter}}\)          & 0,7778          & 0,6775          & \textbf{0,3272}  & 0,6366          & 0,6260          & 0,6020          & 0,6465          & 0,6933          & 0,7261          \\
EL-sup         & 0,7882          & 0,6763          & 0,2829           & 0,6420          & 0,6636          & 0,5832          & 0,6454          & 0,6650          & 0,6971          \\
\(\text{EL}_{\text{filter}}\text{-sup}\)     & \textbf{0,7698} & \textbf{0,6660} & 0,3176           & \textbf{0,6500} & \textbf{0,6951} & \textbf{0,6154} & \textbf{0,6615} & \textbf{0,6996} & \textbf{0,7334} \\
EL-opp         & 0,8083          & 0,6919          & 0,2691           & 0,5841          & 0,6286          & 0,5589          & 0,6040          & 0,6551          & 0,7062          \\
\(\text{EL}_{\text{filter}}\text{-opp}\)     & 0,8064          & 0,6903          & 0,2705           & 0,5899          & 0,6291          & 0,5810          & 0,6336          & 0,6783          & 0,7210          \\
Qwen-Max GenEX & 0,9450          & 0,8171          & -0,0223          & 0,5296          & 0,5900          & 0,5103          & 0,5669          & 0,4695          & 0,5032          \\
\multicolumn{10}{l}{{QwQ}} \\
GenEX          & 0,9599          & 0,8233          & -0,0639          & 0,4511          & 0,4997          & 0,5176          & 0,5706          & 0,4794          & 0,5166          \\
CoT            & 0,8662          & 0,7515          & 0,1535           & 0,5777          & 0,6004          & 0,5509          & 0,6091          & 0,6500          & 0,6916          \\
\(\text{CoT}_{\text{parser}}\)      & 0,8607          & 0,7248          & 0,2536           & 0,6002          & 0,6346          & 0,5632          & 0,6142          & 0,6533          & 0,6965          \\
EL             & 0,8597          & 0,7220          & 0,2670           & 0,6089          & 0,6443          & 0,5669          & 0,6164          & 0,6663          & 0,7153          \\
\(\text{EL}_{\text{filter}}\)          & 0,8023          & 0,6948          & 0,2884           & 0,6350          & 0,6569          & 0,5822          & 0,6365          & \textbf{0,6998} & 0,7235          \\
EL-sup         & 0,7919          & 0,6875          & 0,2817           & 0,6104          & 0,6564          & 0,5876          & 0,6263          & 0,6873          & 0,7397          \\
\(\text{EL}_{\text{filter}}\text{-sup}\)     & \textbf{0,7709} & \textbf{0,6672} & \textbf{0,3212}  & \textbf{0,6394} & \textbf{0,6830} & \textbf{0,5937} & \textbf{0,6513} & 0,6982          & \textbf{0,7417} \\
EL-opp         & 0,8472          & 0,7396          & 0,1844           & 0,5384          & 0,5883          & 0,5085          & 0,5610          & 0,6450          & 0,6872          \\
\(\text{EL}_{\text{filter}}\text{-opp}\)     & 0,8247          & 0,7321          & 0,1999           & 0,5521          & 0,5998          & 0,5404          & 0,5991          & 0,6498          & 0,6984          \\ \midrule
\multicolumn{10}{l}{{llama as judge}} \\
baseline       & 1,0501          & 0,8665          & 0,1211           & 0,4219          & 0,4731          & 0,4937          & 0,5467          & 0,4449          & 0,4815          \\
HumanEX        & 0,9009          & 0,7915          & 0,1045           & 0,1548          & 0,1734          & 0,2050          & 0,2204          & 0,4223          & 0,4508          \\
V3 GenEX       & 1,0338          & 0,9230          & 0,1000           & 0,2915          & 0,3213          & 0,3047          & 0,3355          & 0,2705          & 0,2963          \\
\multicolumn{10}{l}{{R1}} \\
GenEX          & 1,0584          & 0,9383          & 0,1100           & 0,2563          & 0,2907          & 0,2855          & 0,3138          & 0,2199          & 0,2401          \\
CoT            & 0,9099          & 0,9259          & 0,1245           & 0,4427          & 0,4860          & 0,5160          & 0,5449          & 0,4657          & 0,4623          \\
\(\text{CoT}_{\text{parser}}\)      & 0,8988          & 0,8076          & 0,1285           & 0,4454          & 0,4988          & 0,5187          & 0,5577          & 0,4810          & 0,4973          \\
EL             & 0,8860          & 0,7998          & 0,1463           & 0,4539          & 0,5101          & 0,5189          & 0,5680          & 0,4865          & 0,5150          \\
\(\text{EL}_{\text{filter}}\)          & \textbf{0,8602} & 0,7948          & 0,1749           & 0,4888          & 0,5567          & 0,5371          & 0,5736          & 0,5136          & 0,5486          \\
EL-sup         & 0,8909          & 0,7787          & 0,1572           & 0,4861          & 0,5424          & 0,5204          & 0,5750          & 0,4995          & 0,5303          \\
\(\text{EL}_{\text{filter}}\text{-sup}\)     & 0,8760          & \textbf{0,7678} & \textbf{0,2116}  & \textbf{0,5106} & \textbf{0,5626} & \textbf{0,5634} & \textbf{0,6002} & \textbf{0,5296} & \textbf{0,5715} \\
EL-opp         & 0,9071          & 0,8188          & 0,0781           & 0,3558          & 0,4177          & 0,4153          & 0,4317          & 0,4007          & 0,4264          \\
\(\text{EL}_{\text{filter}}\text{-opp}\)     & 0,9025          & 0,8028          & 0,0804           & 0,3572          & 0,4416          & 0,4270          & 0,4459          & 0,4168          & 0,4471          \\
Qwen-Max GenEX & 1,0201          & 0,8949          & 0,0901           & 0,2970          & 0,3198          & 0,3968          & 0,4268          & 0,2802          & 0,3037          \\
\multicolumn{10}{l}{{QwQ}} \\
GenEX          & 1,0606          & 0,9367          & 0,0943           & 0,2855          & 0,3322          & 0,3281          & 0,3612          & 0,1836          & 0,2061          \\
CoT            & 0,8941          & 0,8005          & 0,1519           & 0,4468          & 0,4908          & 0,5282          & 0,5621          & 0,4399          & 0,4637          \\
\(\text{CoT}_{\text{parser}}\)      & 0,8904          & 0,7975          & 0,1580           & 0,4614          & 0,5121          & 0,5378          & 0,5739          & 0,4422          & 0,4686          \\
EL             & 0,8902          & 0,7851          & 0,1612           & 0,4620          & 0,5216          & 0,5469          & 0,6043          & 0,4793          & 0,5093          \\
\(\text{EL}_{\text{filter}}\)          & 0,8835          & 0,7822          & 0,1666           & \textbf{0,5439} & \textbf{0,6069} & 0,5646          & 0,6072          & 0,4908          & 0,5282          \\
EL-sup         & 0,8819          & 0,7831          & 0,1627           & 0,4880          & 0,5440          & 0,5554          & 0,6051          & 0,4846          & 0,5186          \\
\(\text{EL}_{\text{filter}}\text{-sup}\)     & \textbf{0,8442} & \textbf{0,7468} & \textbf{0,2385}  & 0,5317          & 0,5937          & \textbf{0,5751} & \textbf{0,6252} & \textbf{0,5413} & \textbf{0,5644} \\
EL-opp         & 0,9200          & 0,8167          & 0,1125           & 0,3887          & 0,4335          & 0,3941          & 0,4389          & 0,4057          & 0,4090          \\
\(\text{EL}_{\text{filter}}\text{-opp}\)     & 0,9150          & 0,8101          & 0,1271           & 0,3925          & 0,4362          & 0,4297          & 0,4819          & 0,4301          & 0,4407          \\ \midrule
\multicolumn{10}{l}{{mistral as judge}} \\
baseline       & 1,3337          & 1,1461          & -1,0778          & 0,0644          & 0,1059          & 0,4978          & 0,5251          & 0,4661          & 0,4937          \\
HumanEX        & 1,2414          & 1,0864          & -0,7425          & 0,3922          & 0,4106          & 0,4801          & 0,5276          & 0,5903          & 0,6271          \\
V3 GenEX       & 1,3749          & 1,2283          & -1,1310          & 0,3494          & 0,3699          & 0,3812          & 0,4392          & 0,4851          & 0,5132          \\
\multicolumn{10}{l}{{R1}} \\
GenEX          & 1,2964          & 1,1349          & -0,8865          & 0,3187          & 0,3401          & 0,4762          & 0,5287          & 0,4352          & 0,4661          \\
CoT            & 1,0905          & 0,9563          & -0,2716          & 0,5558          & 0,5803          & 0,4828          & 0,5317          & 0,6365          & 0,6949          \\
\(\text{CoT}_{\text{parser}}\)      & 1,0790          & 0,9371          & -0,2219          & 0,5617          & 0,5982          & 0,4855          & 0,5497          & 0,6723          & 0,7038          \\
EL             & 1,0661          & 0,9189          & -0,2129          & 0,5690          & 0,6041          & 0,4986          & 0,5508          & 0,6674          & 0,7098          \\
\(\text{EL}_{\text{filter}}\)          & 1,0470          & 0,9024          & -0,1758          & 0,6205          & 0,6534          & \textbf{0,5160} & 0,5688          & 0,6777          & \textbf{0,7322} \\
EL-sup         & 1,0543          & 0,9099          & -0,1948          & 0,5702          & 0,6047          & 0,5037          & 0,5537          & 0,6762          & 0,7163          \\
\(\text{EL}_{\text{filter}}\text{-sup}\)     & \textbf{1,0172} & \textbf{0,8789} & \textbf{-0,1326} & \textbf{0,6682} & \textbf{0,7053} & 0,5106          & \textbf{0,5768} & \textbf{0,6939} & 0,7254          \\
EL-opp         & 1,1682          & 1,1371          & -0,6817          & 0,4306          & 0,4477          & 0,4547          & 0,5086          & 0,6259          & 0,6831          \\
\(\text{EL}_{\text{filter}}\text{-opp}\)     & 1,1097          & 1,0357          & -0,4609          & 0,5219          & 0,5508          & 0,4667          & 0,5357          & 0,6403          & 0,6958          \\
Qwen-Max GenEX & 1,2137          & 1,0645          & -0,6225          & 0,3974          & 0,4171          & 0,5246          & 0,5731          & 0,5322          & 0,5677          \\
\multicolumn{10}{l}{{QwQ}} \\
GenEX          & 1,3583          & 1,1931          & -1,0457          & 0,3805          & 0,4012          & 0,4051          & 0,4492          & 0,4056          & 0,4331          \\
CoT            & 1,1141          & 1,0037          & -0,3866          & 0,5220          & 0,5418          & 0,5375          & 0,5970          & 0,6413          & 0,6689          \\
\(\text{CoT}_{\text{parser}}\)      & 1,1063          & 0,9781          & -0,3289          & 0,5230          & 0,5488          & 0,5468          & 0,6043          & 0,6499          & 0,6765          \\
EL             & 1,0551          & 0,9755          & -0,2129          & 0,5379          & 0,5686          & 0,5473          & 0,6080          & 0,6600          & 0,6862          \\
\(\text{EL}_{\text{filter}}\)          & 1,0349          & 0,9120          & -0,1833          & 0,5676          & 0,6011          & 0,5793          & \textbf{0,6465} & 0,6822          & 0,7177          \\
EL-sup         & 1,0485          & 0,9147          & -0,1979          & 0,5493          & 0,5794          & 0,5671          & 0,6171          & 0,6712          & 0,6961          \\
\(\text{EL}_{\text{filter}}\text{-sup}\)     & \textbf{1,0188} & \textbf{0,9024} & \textbf{-0,1527} & \textbf{0,6254} & \textbf{0,6596} & \textbf{0,5927} & 0,6384          & \textbf{0,6925} & \textbf{0,7307} \\
EL-opp         & 1,1618          & 1,1472          & -0,6225          & 0,4941          & 0,5082          & 0,4989          & 0,5115          & 0,5974          & 0,6297          \\
\(\text{EL}_{\text{filter}}\text{-opp}\)     & 1,1383          & 1,0741          & -0,5864          & 0,5141          & 0,5191          & 0,5043          & 0,5328          & 0,6230          & 0,6424    \\
\bottomrule
\end{tabular}}
\caption{All HLV evaluation results on SIQA dataset.}
\label{tab:applast-HLV-siqa}
\end{table*}

\begin{table*}[t]
\centering
\resizebox{0.65\linewidth}{!}{
\begin{tabular}{lccccccccc}
\toprule
\multicolumn{1}{c}{\multirow{2}{*}{Settings/Metrics}} & \multicolumn{3}{c}{Score}                    & \multicolumn{2}{c}{Rank-rank}     & \multicolumn{2}{c}{Rank-logits}   & \multicolumn{2}{c}{Rank-score}    \\ \cmidrule(lr){2-4} \cmidrule(lr){5-6} \cmidrule(lr){7-8} \cmidrule(lr){9-10} 
\multicolumn{1}{c}{}                                  & RMSE ↓            & MAE ↓           & $R^2$ ↑           & $\tau$ ↑          & $\rho$ ↑  & $\tau$ ↑           & $\rho$ ↑  & $\tau$ ↑          & $\rho$ ↑ \\ \midrule
\multicolumn{10}{l}{{qwen as judge}} \\
baseline       & 0,9101          & 0,7417          & 0,4255          & 0,5395          & 0,6283          & 0,4509          & 0,5692          & 0,5953          & 0,6332          \\
HumanEX        & 0,9209          & 0,7536          & 0,4205          & 0,4507          & 0,5225          & 0,4900          & 0,5754          & 0,5824          & 0,6484          \\
V3 GenEX       & 0,9761          & 0,8453          & 0,3275          & 0,5461          & 0,6347          & 0,4296          & 0,5496          & 0,5262          & 0,5481          \\
\multicolumn{10}{l}{{R1}} \\
GenEX          & 0,9757          & 0,8004          & 0,3492          & 0,5576          & 0,6383          & 0,4571          & 0,5692          & 0,5708          & 0,5946          \\
CoT            & 0,8856          & 0,7317          & 0,3992          & 0,5050          & 0,6004          & 0,4618          & 0,5752          & 0,6077          & 0,6461          \\
\(\text{CoT}_{\text{parser}}\)      & 0,8849          & 0,7298          & 0,4428          & 0,5716          & 0,6419          & 0,4680          & 0,5780          & 0,6112          & 0,6738          \\
EL             & 0,8845          & 0,7298          & 0,4554          & 0,5957          & 0,6492          & 0,4786          & 0,5830          & 0,6275          & 0,6852          \\
\(\text{EL}_{\text{filter}}\)          & 0,8649          & 0,7127          & {0,4887} & 0,6104          & 0,6770          & 0,4998          & 0,6110          & 0,6319          & 0,7001          \\
EL-sup         & 0,8704          & 0,7153          & 0,4844          & 0,6094          & 0,6736          & 0,4883          & 0,5928          & 0,6301          & 0,6997          \\
\(\text{EL}_{\text{filter}}\text{-sup}\)     & \textbf{0,8646} & \textbf{0,6956} & \textbf{0,4937} & \textbf{0,6114} & \textbf{0,6790} & \textbf{0,5152} & \textbf{0,6180} & \textbf{0,6605} & \textbf{0,7313} \\
EL-opp         & 0,9722          & 0,8108          & 0,3265          & 0,5330          & 0,6126          & 0,4556          & 0,5453          & 0,5853          & 0,6728          \\
\(\text{EL}_{\text{filter}}\text{-opp}\)     & 0,9584          & 0,8024          & 0,3276          & 0,5499          & 0,6163          & 0,4653          & 0,5554          & 0,5953          & 0,6832          \\
Qwen-Max GenEX & 0,9830          & 0,8387          & 0,3838          & 0,5599          & 0,6293          & 0,4512          & 0,5399          & 0,4984          & 0,6374          \\
\multicolumn{10}{l}{{QwQ}} \\
GenEX          & 0,9607          & 0,8147          & 0,3998          & 0,5416          & 0,6349          & 0,4599          & 0,5553          & 0,5533          & 0,6493          \\
CoT            & 0,9048          & 0,7498          & 0,4057          & 0,5884          & 0,6582          & 0,4696          & 0,5456          & 0,5869          & 0,6676          \\
\(\text{CoT}_{\text{parser}}\)      & 0,9006          & 0,7326          & 0,4329          & 0,6253          & 0,6734          & 0,4839          & 0,5722          & 0,6087          & 0,6710          \\
EL             & 0,8882          & 0,7317          & 0,4357          & 0,6270          & 0,6966          & 0,4921          & 0,5849          & 0,6243          & 0,6844          \\
\(\text{EL}_{\text{filter}}\)          & 0,8786          & 0,7203          & 0,4416          & 0,6344          & 0,7056          & 0,4997          & 0,5948          & {0,6372} & 0,7001          \\
EL-sup         & 0,8880          & 0,7211          & 0,4377          & 0,6273          & 0,7019          & 0,4993          & 0,5977          & 0,6323          & 0,6938          \\
\(\text{EL}_{\text{filter}}\text{-sup}\)     & \textbf{0,8787} & \textbf{0,7197} & \textbf{0,4541} & \textbf{0,6378} & \textbf{0,7109} & \textbf{0,5126} & \textbf{0,5999} & \textbf{0,6432} & \textbf{0,7068} \\
EL-opp         & 0,9407          & 0,8821          & 0,3319          & 0,5480          & 0,6315          & 0,4553          & 0,5472          & 0,5967          & 0,6546          \\
\(\text{EL}_{\text{filter}}\text{-opp}\)     & 0,9305          & 0,8000          & 0,3879          & 0,5623          & 0,6493          & 0,4625          & 0,5568          & 0,6036          & 0,6799          \\ \midrule
\multicolumn{10}{l}{{llama as judge}} \\
baseline       & 1,1724          & 1,1788          & 0,1980          & 0,4809          & 0,5707          & 0,3690          & 0,4313          & 0,4123          & 0,4562          \\
HumanEX        & 1,0798          & 0,9270          & 0,2416          & 0,4663          & 0,5452          & 0,2889          & 0,3551          & 0,3823          & 0,4358          \\
V3 GenEX       & 1,2123          & 1,0354          & 0,0417          & 0,3650          & 0,4337          & 0,2852          & 0,3365          & 0,3236          & 0,3807          \\
\multicolumn{10}{l}{{R1}} \\
GenEX          & 1,2731          & 1,1097          & -0,0546         & 0,4332          & 0,5210          & 0,3168          & 0,3789          & 0,2921          & 0,3525          \\
CoT            & 1,1958          & 1,1164          & 0,1308          & 0,4655          & 0,5552          & 0,3603          & 0,4233          & 0,3723          & 0,4173          \\
\(\text{CoT}_{\text{parser}}\)      & 1,1775          & 1,0275          & 0,1408          & 0,4685          & 0,5500          & 0,3612          & 0,4393          & 0,4076          & 0,4602          \\
EL             & 1,1605          & 1,0091          & 0,1609          & 0,4844          & 0,5670          & 0,3717          & 0,4418          & 0,4161          & 0,4603          \\
\(\text{EL}_{\text{filter}}\)          & {1,1359} & 1,0727          & \textbf{0,1853} & 0,5117          & 0,5930          & 0,3732          & 0,4499          & 0,4223          & 0,4762          \\
EL-sup         & 1,1429          & 1,0051          & 0,1685          & 0,4924          & 0,5865          & 0,3742          & 0,4467          & 0,4182          & 0,4844          \\
\(\text{EL}_{\text{filter}}\text{-sup}\)     & \textbf{1,1302} & \textbf{1,0045} & {0,1780} & \textbf{0,5176} & \textbf{0,6030} & \textbf{0,3782} & \textbf{0,4534} & \textbf{0,4265} & \textbf{0,4867} \\
EL-opp         & 1,2152          & 1,0848          & 0,0839          & 0,4367          & 0,5072          & 0,3467          & 0,4146          & 0,3434          & 0,3950          \\
\(\text{EL}_{\text{filter}}\text{-opp}\)     & 1,2144          & 1,0801          & 0,0722          & 0,4494          & 0,5288          & 0,3544          & 0,4229          & 0,3578          & 0,4089          \\
Qwen-Max GenEX & 1,1995          & 1,0387          & 0,0515          & 0,4537          & 0,5391          & 0,3393          & 0,4103          & 0,3202          & 0,3675          \\
\multicolumn{10}{l}{{QwQ}} \\
GenEX          & 1,2349          & 1,0812          & 0,0088          & 0,4610          & 0,5275          & 0,2888          & 0,3469          & 0,2764          & 0,3261          \\
CoT            & 1,3606          & 1,0361          & 0,1260          & 0,4691          & 0,5564          & 0,3609          & 0,4438          & 0,3763          & 0,4483          \\
\(\text{CoT}_{\text{parser}}\)      & 1,1565          & 1,0135          & 0,1285          & 0,4781          & 0,5636          & 0,3665          & 0,4451          & 0,3993          & 0,4727          \\
EL             & 1,1533          & 1,0127          & 0,1259          & 0,4844          & 0,5704          & 0,3669          & 0,4479          & 0,4158          & 0,4800          \\
\(\text{EL}_{\text{filter}}\)          & 1,1499          & 1,0059          & 0,1533          & {0,5032} & {0,5858} & 0,3727          & 0,4525          & 0,4509          & 0,5097          \\
EL-sup         & 1,1477          & 1,0127          & 0,1532          & 0,5005          & 0,5765          & 0,3685          & 0,4495          & 0,4480          & 0,5086          \\
\(\text{EL}_{\text{filter}}\text{-sup}\)     & \textbf{1,1317} & \textbf{0,9921} & \textbf{0,1658} & \textbf{0,5238} & \textbf{0,5988} & \textbf{0,3733} & \textbf{0,4649} & \textbf{0,4631} & \textbf{0,5311} \\
EL-opp         & 1,1678          & 1,0695          & 0,1104          & 0,4132          & 0,5068          & 0,3476          & 0,4124          & 0,3148          & 0,3698          \\
\(\text{EL}_{\text{filter}}\text{-opp}\)     & 1,1500          & 1,0454          & 0,1254          & 0,4399          & 0,5403          & 0,3553          & 0,4285          & 0,3216          & 0,4306          \\ \midrule
\multicolumn{10}{l}{{mistral as judge}} \\
baseline       & 1,5770          & 1,2886          & -0,7480         & 0,3011          & 0,3429          & 0,3768          & 0,4466          & 0,4024          & 0,4416          \\
HumanEX        & 1,2543          & 1,0163          & -0,0984         & 0,2892          & 0,3117          & 0,3634          & 0,4328          & 0,5096          & 0,5692          \\
V3 GenEX       & 1,6096          & 1,3165          & -0,7930         & 0,3751          & 0,4271          & 0,3058          & 0,3739          & 0,4050          & 0,4483          \\
\multicolumn{10}{l}{{R1}} \\
GenEX          & 1,7442          & 1,4374          & -1,0796         & 0,3607          & 0,4160          & 0,3585          & 0,4274          & 0,3245          & 0,3635          \\
CoT            & 1,2010          & 0,9702          & -0,0156         & 0,4621          & 0,5134          & 0,3761          & 0,4454          & 0,5890          & 0,6643          \\
\(\text{CoT}_{\text{parser}}\)      & 1,2000          & 0,9606          & -0,0081         & 0,4658          & 0,5227          & 0,3806          & 0,4539          & 0,5988          & 0,6704          \\
EL             & 1,1539          & 0,9510          & 0,0549          & 0,4752          & 0,5262          & 0,3816          & 0,4564          & 0,6055          & 0,6770          \\
\(\text{EL}_{\text{filter}}\)          & 1,1461          & 0,9354          & 0,0727          & 0,5148          & 0,5678          & {0,3860} & 0,4572          & \textbf{0,6139} & \textbf{0,6836} \\
EL-sup         & 1,1491          & 0,9437          & 0,0629          & 0,4782          & 0,5303          & 0,3827          & 0,4566          & 0,6118          & 0,6786          \\
\(\text{EL}_{\text{filter}}\text{-sup}\)     & \textbf{1,1398} & \textbf{0,9229} & \textbf{0,0737} & \textbf{0,5298} & \textbf{0,5882} & \textbf{0,3981} & \textbf{0,4685} & {0,6133} & 0,6811          \\
EL-opp         & 1,2232          & 0,9773          & -0,0522         & 0,4486          & 0,4925          & 0,3468          & 0,4015          & 0,5449          & 0,5761          \\
\(\text{EL}_{\text{filter}}\text{-opp}\)     & 1,2057          & 0,9667          & -0,0296         & 0,4586          & 0,5043          & 0,3523          & 0,4333          & 0,5730          & 0,6031          \\
Qwen-Max GenEX & 1,4453          & 1,1642          & -0,4438         & 0,4025          & 0,4581          & 0,3527          & 0,4122          & 0,4599          & 0,5177          \\
\multicolumn{10}{l}{{QwQ}} \\
GenEX          & 1,6740          & 1,3514          & -0,8915         & 0,3523          & 0,3962          & 0,3549          & 0,4270          & 0,3812          & 0,4185          \\
CoT            & 1,1811          & 0,9594          & 0,0101          & 0,4875          & 0,5471          & 0,3604          & 0,4356          & 0,5876          & 0,6484          \\
\(\text{CoT}_{\text{parser}}\)      & 1,1674          & 0,9456          & 0,0147          & 0,4920          & 0,5448          & 0,3661          & 0,4410          & 0,5914          & 0,6575          \\
EL             & 1,1624          & 0,9443          & 0,0723          & 0,4945          & 0,5625          & 0,3707          & 0,4469          & 0,6070          & 0,6700          \\
\(\text{EL}_{\text{filter}}\)          & 1,1306          & 0,9235          & 0,0869          & \textbf{0,5370} & 0,5915          & 0,3755          & {0,4561} & 0,6146          & 0,6826          \\
EL-sup         & 1,1414          & 0,9232          & 0,0751          & 0,5163          & 0,5727          & 0,3711          & 0,4472          & 0,6171          & 0,6830          \\
\(\text{EL}_{\text{filter}}\text{-sup}\)     & \textbf{1,1211} & \textbf{0,9056} & \textbf{0,1132} & {0,5361} & \textbf{0,5956} & \textbf{0,3917} & \textbf{0,4638} & \textbf{0,6287} & \textbf{0,6958} \\
EL-opp         & 1,2116          & 0,9859          & -0,0387         & 0,4322          & 0,4827          & 0,3398          & 0,4127          & 0,5688          & 0,6302          \\
\(\text{EL}_{\text{filter}}\text{-opp}\)     & 1,1950          & 0,9661          & -0,0139         & 0,4419          & 0,5152          & 0,3654          & 0,4282          & 0,5911          & 0,6489       \\
\bottomrule
\end{tabular}}
\caption{All HLV evaluation results on CQA dataset.}
\label{tab:applast-HLV-cqa}
\end{table*}

\section{Use of AI Assistants}

The authors acknowledge the use of ChatGPT solely for correcting grammatical errors, enhancing the coherence of the final manuscript.

\end{document}